\documentclass[twocolumn,floatfix,superscriptaddress,longbibliography,notitlepage
]{revtex4-2} 

\usepackage{amsmath,dsfont}
\usepackage{graphicx}
\usepackage{dcolumn}
\usepackage{bm}
\usepackage{color}
\usepackage{hyperref}
\usepackage{svg}
\usepackage{multirow}
\usepackage[super]{nth}
\usepackage{titlesec}
\titlespacing\subsubsection{1ex}{1.0ex plus 1ex minus -.2ex}{\parskip}

\graphicspath{{main_fig/}{supplementary_fig/}{architecture_fig/}}

\hypersetup{
    colorlinks=true,
    linkcolor=black,
    citecolor=black,
    filecolor=magenta,      
    urlcolor=cyan,
    pdftitle={The Backpropagation Algorithm Implemented on Spiking Neuromorphic Hardware},
    pdfsubject={cs.NC},
    pdfauthor={Alpha Renner, Forrest Sheldon, Anatoly Zlotnik, Louis Tao, Andrew Sornborger},
    pdfkeywords={Backpropagation, SNN, Spiking Neural Network, Neuromorphic Hardware},
}

\newcommand{\gpuSinglepowerdynamic}{19000}
\newcommand{\gpuSinglepowertotal}{34000}
\newcommand{\gpuSinglelatencypertrial}{1.0}
\newcommand{\gpuSingleenergypertrialdynamic}{19.0}
\newcommand{\gpuSingleedp}{19.0}
\newcommand{\gpuBatchTenpowerdynamic}{19000}
\newcommand{\gpuBatchTenpowertotal}{34000}
\newcommand{\gpuBatchTenlatencypertrial}{0.1}
\newcommand{\gpuBatchTenenergypertrialdynamic}{1.9}
\newcommand{\gpuBatchTenedp}{0.19}
\newcommand{\gpuBatchHundredpowerdynamic}{19000}
\newcommand{\gpuBatchHundredpowertotal}{34000}
\newcommand{\gpuBatchHundredlatencypertrial}{0.01}
\newcommand{\gpuBatchHundredenergypertrialdynamic}{0.19}
\newcommand{\gpuBatchHundrededp}{0.0019}

\newcommand{\Loihilearningoffpowerlakemontdynamic}{22.2}

\newcommand{\Loihilearningoffpowerlakemontstatic}{0.125}
\newcommand{\Loihilearningoffpowerlakemonttotal}{22.3}
\newcommand{\Loihilearningoffpowerdynamic}{77.3}
\newcommand{\Loihilearningoffpowercorestatic}{34.1}
\newcommand{\Loihilearningoffpowercoredynamic}{55.1}

\newcommand{\Loihilearningoffpowercoretotal}{89.2}

\newcommand{\Loihilearningoffpowertotal}{1031}
\newcommand{\Loihilearningoffpowerstatic}{954}

\newcommand{\Loihilearningofflatencypertrial}{0.37}

\newcommand{\LoihilearningoffenergypertrialBoardDynamic}{0.0286}
\newcommand{\LoihilearningoffenergypertrialLoihiDynamic}{0.0204}
\newcommand{\LoihilearningoffenergypertrialLakemontDynamic}{0.00822}

\newcommand{\LoihilearningoffedpLakemontDynamic}{0.00304}
\newcommand{\LoihilearningoffedpLoihiDynamic}{0.00755}
\newcommand{\LoihilearningoffedpBoardDynamic}{0.0106}

\newcommand{\Loihitrainingdownsampledpowertotal}{1128}
\newcommand{\Loihitrainingdownsampledpowerstatic}{916}

\newcommand{\Loihitrainingdownsampledpowerlakemontstatic}{0.119}

\newcommand{\Loihitrainingdownsampledpowerlakemontdynamic}{19.0}

\newcommand{\Loihitrainingdownsampledpowerlakemonttotal}{19.1}
\newcommand{\Loihitrainingdownsampledpowerdynamic}{212}
\newcommand{\Loihitrainingdownsampledpowercorestatic}{35.7}
\newcommand{\Loihitrainingdownsampledpowercoredynamic}{193}

\newcommand{\Loihitrainingdownsampledpowercoretotal}{229}

\newcommand{\Loihitrainingdownsampledlatencypertrial}{0.731}

\newcommand{\LoihitrainingdownsampledenergypertrialBoardDynamic}{0.155}
\newcommand{\LoihitrainingdownsampledenergypertrialLoihiDynamic}{0.141}
\newcommand{\LoihitrainingdownsampledenergypertrialLakemontDynamic}{0.0139}

\newcommand{\LoihitrainingdownsamplededpLakemontDynamic}{0.0101}
\newcommand{\LoihitrainingdownsamplededpLoihiDynamic}{0.103}
\newcommand{\LoihitrainingdownsamplededpBoardDynamic}{0.113}

\newcommand{\Loihitrainingsmallerpowercorestatic}{40.0}

\newcommand{\Loihitrainingsmallerpowercoredynamic}{399}

\newcommand{\Loihitrainingsmallerpowercoretotal}{439}

\newcommand{\Loihitrainingsmallerpowerstatic}{914}
\newcommand{\Loihitrainingsmallerpowerdynamic}{418}

\newcommand{\Loihitrainingsmallerpowerlakemontdynamic}{18.9}

\newcommand{\Loihitrainingsmallerpowerlakemontstatic}{0.133}

\newcommand{\Loihitrainingsmallerpowerlakemonttotal}{19.0}
\newcommand{\Loihitrainingsmallerpowertotal}{1332}

\newcommand{\Loihitrainingsmallerlatencypertrial}{1.18}

\newcommand{\LoihitrainingsmallerenergypertrialBoardDynamic}{0.494}
\newcommand{\LoihitrainingsmallerenergypertrialLoihiDynamic}{0.471}
\newcommand{\LoihitrainingsmallerenergypertrialLakemontDynamic}{0.0223}

\newcommand{\LoihitrainingsmalleredpLakemontDynamic}{0.0264}
\newcommand{\LoihitrainingsmalleredpLoihiDynamic}{0.557}
\newcommand{\LoihitrainingsmalleredpBoardDynamic}{0.583}

\newcommand{\Loihitrainingstartpowertotal}{1328}
\newcommand{\Loihitrainingstartpowerstatic}{912}
\newcommand{\Loihitrainingstartpowerdynamic}{416}

\newcommand{\Loihitrainingstartpowerlakemontdynamic}{18.5}

\newcommand{\Loihitrainingstartpowerlakemontstatic}{0.137}

\newcommand{\Loihitrainingstartpowerlakemonttotal}{18.6}
\newcommand{\Loihitrainingstartpowercoredynamic}{398}

\newcommand{\Loihitrainingstartpowercorestatic}{37.6}
\newcommand{\Loihitrainingstartpowercoretotal}{435}

\newcommand{\Loihitrainingstartlatencypertrial}{1.49}

\newcommand{\LoihitrainingstartenergypertrialBoardDynamic}{0.619}
\newcommand{\LoihitrainingstartenergypertrialLoihiDynamic}{0.591}
\newcommand{\LoihitrainingstartenergypertrialLakemontDynamic}{0.0275}

\newcommand{\LoihitrainingstartedpLakemontDynamic}{0.0409}
\newcommand{\LoihitrainingstartedpLoihiDynamic}{0.88}
\newcommand{\LoihitrainingstartedpBoardDynamic}{0.92}

\newcommand{\Loihitrainingendpowerdynamic}{418}
\newcommand{\Loihitrainingendpowerstatic}{914}

\newcommand{\Loihitrainingendpowercorestatic}{41.1}
\newcommand{\Loihitrainingendpowercoredynamic}{399}

\newcommand{\Loihitrainingendpowercoretotal}{440}

\newcommand{\Loihitrainingendpowerlakemontdynamic}{18.5}
\newcommand{\Loihitrainingendpowerlakemontstatic}{0.15}

\newcommand{\Loihitrainingendpowerlakemonttotal}{18.7}
\newcommand{\Loihitrainingendpowertotal}{1332}

\newcommand{\LoihitrainingendnumCores}{81}
\newcommand{\Loihitrainingendlatencypertrial}{1.48}

\newcommand{\LoihitrainingendenergypertrialLoihi}{0.653}

\newcommand{\LoihitrainingendenergypertrialBoardDynamic}{0.62}
\newcommand{\LoihitrainingendenergypertrialLoihiDynamic}{0.592}
\newcommand{\LoihitrainingendenergypertrialLakemontDynamic}{0.0275}

\newcommand{\LoihitrainingendedpLakemontDynamic}{0.0408}
\newcommand{\LoihitrainingendedpLoihiDynamic}{0.878}
\newcommand{\LoihitrainingendedpBoardDynamic}{0.919}

\newcommand{\Loihiinferencepowertotal}{1019}

\newcommand{\Loihiinferencepowercoredynamic}{14.7}

\newcommand{\Loihiinferencepowercorestatic}{39.2}
\newcommand{\Loihiinferencepowercoretotal}{53.9}
\newcommand{\Loihiinferencepowerdynamic}{38.2}

\newcommand{\Loihiinferencepowerlakemontdynamic}{23.4}
\newcommand{\Loihiinferencepowerlakemontstatic}{0.143}
\newcommand{\Loihiinferencepowerlakemonttotal}{23.6}
\newcommand{\Loihiinferencepowerstatic}{980}

\newcommand{\Loihiinferencelatencypertrial}{0.169}

\newcommand{\LoihiinferenceenergypertrialBoardDynamic}{0.00645}
\newcommand{\LoihiinferenceenergypertrialLoihiDynamic}{0.00249}
\newcommand{\LoihiinferenceenergypertrialLakemontDynamic}{0.00396}

\newcommand{\LoihiinferenceedpLakemontDynamic}{0.000669}
\newcommand{\LoihiinferenceedpLoihiDynamic}{0.000421}
\newcommand{\LoihiinferenceedpBoardDynamic}{0.00109}

\newcommand{\bestaccuracy}{96.2}
\newcommand{\finalaccuracy}{95.7}
\newcommand{\finalepoch}{60}

\begin{document}


\title{The Backpropagation Algorithm Implemented on Spiking Neuromorphic Hardware}



\author{Alpha Renner}  \email{alpren@ini.uzh.ch}
 \affiliation{Institute of Neuroinformatics, University of Zurich and ETH Zurich, Zurich, Switzerland, 8057}

\author{Forrest Sheldon} \email{fsheldon@lanl.gov}
 \affiliation{Physics of Condensed Matter \& Complex Systems (T-4), Los Alamos National Laboratory, Los Alamos, New Mexico, USA, 87545}

\author{Anatoly Zlotnik}  \email{azlotnik@lanl.gov}
\affiliation{Applied Mathematics \& Plasma Physics (T-5), Los Alamos National Laboratory, Los Alamos, New Mexico, USA, 87545}

\author{Louis Tao}  \email{taolt@mail.cbi.pku.edu.cn}
\affiliation{Center for Bioinformatics, National Laboratory of Protein Engineering and Plant Genetic Engineering, School of Life Sciences, Peking University, Beijing, China, 100871}
\affiliation{Center for Quantitative Biology, Academy for Advanced Interdisciplinary Studies, Peking University, Beijing, China, 100871}

\author{Andrew Sornborger}  \email{sornborg@lanl.gov}
 \thanks{Corresponding Author}
\affiliation{Information Sciences (CCS-3), Los Alamos National Laboratory, Los Alamos, New Mexico, USA, 87545}

\date{\today}

\begin{abstract}
\ifdefined\linenumbers \linenumbers \fi 

\noindent
The capabilities of natural neural systems have inspired new generations of machine learning algorithms as well as neuromorphic very large-scale integrated (VLSI) circuits capable of fast, low-power information processing.
%
However, it has been argued that most modern machine learning algorithms are not neurophysiologically plausible.
%
In particular, the workhorse of modern deep learning, the backpropagation algorithm, has proven difficult to translate to neuromorphic hardware. 
In this study, we present a neuromorphic, spiking backpropagation algorithm based on synfire-gated dynamical information coordination and processing, implemented on Intel's Loihi neuromorphic research processor.
%
%
We demonstrate a proof-of-principle three-layer circuit that learns to classify digits from the MNIST dataset.
To our knowledge, this is the first work to show a Spiking Neural Network (SNN) implementation of the backpropagation algorithm that is fully on-chip, without a computer in the loop.
It is competitive in accuracy with off-chip trained SNNs and achieves an energy-delay product suitable for edge computing.
%
%
This implementation shows a path for using in-memory, massively parallel neuromorphic processors for low-power, low-latency implementation of modern deep learning applications.

\end{abstract}

\pacs{07.05.Mh, 07.05.Pj, 87.19.lr, 87.19.ls, 87.19.lv, 87.85.Wc}  

\ifdefined\linenumbers \begin{nolinenumbers} \fi 
\maketitle
\ifdefined\linenumbers \end{nolinenumbers} \fi 
\section{Introduction}
\label{sec:main}
\subsubsection{Dynamical Coordination in Cognitive Processes.}
Spike-based learning in plastic neuronal networks is playing increasingly key roles in both theoretical neuroscience and neuromorphic computing. The brain learns in part by modifying the synaptic strengths between neurons and neuronal populations. While specific synaptic plasticity or neuromodulatory mechanisms may vary in different brain regions, it is becoming clear that a significant level of dynamical coordination between disparate neuronal populations must exist, even within an individual neural circuit \cite{RoelfsemaHoltmaat2018}.
Classically, backpropagation (BP, and other learning algorithms) \cite{linnainmaa1970representation,werbos1974beyond,rumelhart_1985} has been essential for supervised learning in artificial neural networks (ANNs). Although the question of whether or not BP operates in the brain is still an outstanding issue \cite{lillicrap2020backpropagation}, BP does solve the problem of how a global objective function can be related to local synaptic modification in a network. It seems clear, however, that if BP is implemented in the brain, or if one wishes to implement BP in a neuromorphic circuit, some amount of dynamical information coordination is necessary to propagate the correct information to the correct location such that appropriate local synaptic modification may take place to enable learning.

\subsubsection{The Grand Challenge of Spiking Backpropagation (sBP)}
There has been a rapid growth of interest in the reformulation of classical algorithms for learning, optimization, and control using event-based information-processing mechanisms.  Such spiking neural networks (SNNs) are inspired by the function of biophysiological neural systems \cite{pmid26906502}, i.e.,  neuromorphic computing \cite{mead1990neuromorphic}.  The trend is driven by the advent of flexible computing architectures such as Intel’s neuromorphic research processor, codenamed Loihi, that enable experimentation with such algorithms in hardware \cite{DBLP:journals/micro/DaviesSLCCCDJIJ18}.  There is particular interest in deep learning, which is a central tool in modern machine learning.  Deep learning relies on a layered, feedforward network similar to the early layers of the visual cortex, with threshold nonlinearities at each layer that resemble mean-field approximations of neuronal integrate-and-fire models.  While feedforward networks are readily translated to neuromorphic hardware~\cite{EsserEtAl2016, rueckauer2017conversion,severa2019training}, the far more computationally intensive training of these networks `on chip' has proven elusive as the structure of backpropagation makes the algorithm notoriously difficult to implement in a neural circuit \cite{Grossberg:1988:CLI:54455.54464,Crick1989}. A feasible neural implementation of the backpropagation algorithm has gained renewed scrutiny with the rise of new neuromorphic computational architectures that feature local synaptic plasticity \cite{DBLP:journals/micro/DaviesSLCCCDJIJ18,PainkrasEtAl2012,SchemmelEtAl2010, qiao2015reconfigurable}. Because of the well-known difficulties, neuromorphic systems have relied to date almost entirely on conventional off-chip learning, and used on-chip computing only for inference \cite{EsserEtAl2016,rueckauer2017conversion,severa2019training}.  It has been a long-standing challenge to develop learning systems whose function is realized exclusively using neuromorphic mechanisms.

Backpropagation has been claimed to be biologically implausible or difficult to implement on spiking chips because of several issues:
(a) {\it Weight transport} -- Usually, synapses in biology and on neuromorphic hardware cannot be used bidirectionally, therefore, separate synapses for the forward and backward pass are employed. However, correct credit assignment, i.e. knowing how a weight change affects the error, requires feedback weights to be the same as feedforward weights \cite{grossberg1987competitive,liao2015};
(b) {\it Backwards computation} --  Forward and backward passes implement different computations~\cite{liao2015}. The forward pass requires only weighted summation of the inputs, while the backward pass operates in the opposite direction and additionally takes into account the derivative of the activation function;
(c) {\it Gradient storage} -- Error gradients must be computed and stored separately from activations; 
(d) {\it Differentiability} -- For spiking networks, the issue of non-differentiability of spikes has been discussed and solutions have been proposed~\cite{bohte_2002,pfister_2006,zenke2018superspike,huh2018gradient}; and
(e) {\it Hardware constraints} -- For the case of neuromorphic hardware, there are often constraints on plasticity mechanisms, which allow for adaptation of synaptic weights. On some hardware, no plasticity is offered at all, while in some cases only specific spike-timing dependent plasticity (STDP) rules are allowed. Additionally, in almost all available neuromorphic architectures, information must be local, i.e., information is only shared between neurons that are synaptically connected, in particular, to facilitate parallelization. 

\subsubsection{Previous Attempts at sBP} 
The most commonly used approach to avoiding the above issues is to use neuromorphic hardware only for inference using fixed weights, obtained by training of an identical network offline and off-chip \citep{EsserEtAl2016, rueckauer2017conversion, rasmussen2019nengodl, sengupta_2019, shrestha2018slayer, nair2019mapping, rueckauer2021nxtf}. This has recently achieved state-of-the-art performance \citep{severa2019training}, but it does not make use of neuromorphic hardware's full potential, because offline training consumes significant power. Moreover, to function in most field applications, an inference algorithm should be able to learn adaptively after deployment, e.g., to adjust to a particular speaker in speech recognition, which would enable autonomy and privacy of edge computing devices. So far, only last layer training, without backpropagation, and using variants of the delta rule, has been achieved on spiking hardware \cite{stewart2020chip,dewolf2020nengo, frenkel20180, kim2015640m, buhler20173, park20197, nandakumar2020experimental}. Other on-chip learning approaches use alternatives to backpropagation \cite{frenkel202028, shrestha2021hardware}, bio-inspired non-gradient based methods \cite{imam2020rapid}, or hybrid systems with a conventional computer in the loop \cite{friedmann2016demonstrating, nandakumar2020mixed}.
Several promising alternative approaches for actual on-chip spiking backpropagation have been proposed recently \cite{payvand2020chip, payeur2020burst, bellec2020solution, sacramento2018dendritic}, but have not yet been implemented in hardware.

\begin{figure}[!ht]
  \centering
  \includegraphics[width=0.85\linewidth]{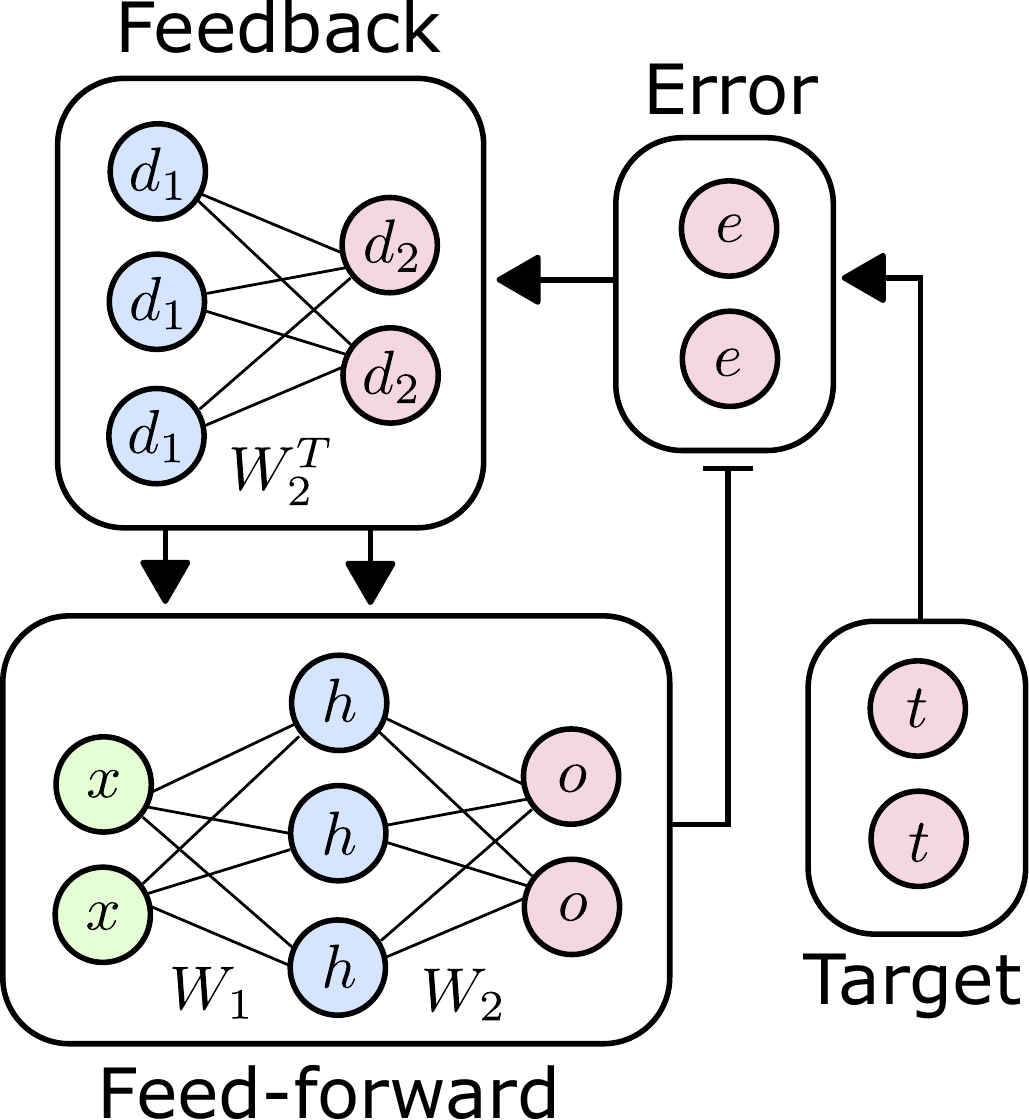}
  \caption{\textbf{Overview of conceptual circuit architecture.} Feedforward activations of input (${x}$), hidden (${h}$) and output (${o}$) layers are calculated by a feedforward module.  Errors (${e} = {t} - {o}$) are calculated from the output and the training signal (${t}$).  Errors are backpropagated through a feedback module with the same weights $W_2$ for synapses between ${h}$ and ${o}$, but in the opposite direction (mathematically expressed as the transpose, $W_2^T$). Local gradients (${d}_{1},\,{d}_{2}$) are gated back into the feedforward circuit at appropriate times to accomplish potentiation or depression of appropriate weights. 
  } \label{fig:Arch}
\end{figure}

To avoid the backwards computation issue (b) and because neuromorphic synapses are not bidirectional, a separate feedback network for the backpropagation of errors has been proposed  \citep{stork1989backpropagation,Zipser1993computational} (see Fig.~\ref{fig:Arch}). This leads to the weight transport problem (a) which has been solved by using symmetric learning rules to maintain weight symmetry \citep{oreilly_1996,Zipser1993computational,kolen_1994} or with the Kolen-Pollack algorithm \citep{kolen_1994}, which leads to symmetric weights automatically. It has also been found that weights do not have to be perfectly symmetric, because backpropagation can still be effective with random feedback weights (random feedback alignment) \citep{lillicrap2016random}, although symmetry in the sign between forward and backward weights matters \citep{Liao:2016:IWS:3016100.3016156}. 

The backwards computation issue (b) and the gradient storage issue (c) have been addressed by approaches that separate the function of the neuron into different compartments and use structures that resemble neuronal dendrites for the computation of backward propagating errors \citep{richards_2019,payeur2020burst,sacramento2018dendritic,lillicrap2020backpropagation}.  The differentiability issue (d) has been circumvented by spiking rate-based approaches \citep{oconnor_2013, rueckauer2017conversion, sengupta_2019, kim2019simple} that use the ReLU activation function as done in ANNs.  The differentiability issue has also been addressed more generally using surrogate gradient methods \cite{bohte_2002,pfister_2006,EsserEtAl2016,zenke2018superspike,huh2018gradient,shrestha2018slayer,stewart2020chip,neftci2019surrogate} and methods that use biologically-inspired STDP and reward modulated STDP mechanisms \citep{izhikevich_2007,sporea_2013, legenstein_2008,fremaux_2015}.
For a review of SNN-based deep learning, see \cite{tavanaei2019deep}. For a review of backpropagation in the brain, see \cite{lillicrap2020backpropagation}.

\subsubsection{Our Contribution}
In this study, we describe a hardware implementation of the backpropagation algorithm that addresses each of the issues (a)--(e) introduced above using a set of mechanisms that have been developed and tested in simulation by the authors during the past decade, synthesized in our recent study \cite{sornborger2019pulse}, and simplified and adapted here to the features and constraints of the Loihi chip. 
These neuronal and network features include propagation of graded information in a circuit composed of neural populations using synfire-gated synfire chains (SGSCs) \cite{SornborgerWangTao,WangSornborgerTao,wang2015fokker,XiaoWangSornborgerTao}, control flow based on the interaction of synfire-chains \cite{WangSornborgerTao}, and regulation of Hebbian learning using synfire-gating \cite{ShaoSornborgerTao,ShaoWangSornborgerTao2018}.  We simplify our previously proposed network architecture \cite{sornborger2019pulse}, and streamline its function.  We demonstrate our approach using a proof-of-principle implementation on Loihi \cite{DBLP:journals/micro/DaviesSLCCCDJIJ18}, and examine the performance of the algorithm for learning and inference of the MNIST test data set \cite{lecun1998gradient}.  The sBP implementation is shown to be competitive in clock time, sparsity, and power consumption with state-of-the-art algorithms for the same tasks.

\section{Results}  \label{sec:results}

\subsubsection{The Binarized sBP model}
We extend our previous architecture \cite{sornborger2019pulse} using several new algorithmic and hardware-specific mechanisms. 
Each unit of the neural network is implemented as a single spiking neuron, using the current-based leaky integrate-and-fire (CUBA) model (see Eq.~\eqref{eq:loihi_eq} in Section~\ref{sec:methods}) that is built into Loihi. The time constants of the CUBA model are set to 1, so that the neurons are memoryless. Rather than using rate coding,  whereby spikes are counted over time, we consider neuron spikes at every algorithmic time step, so we can regard our implementation as a binary neural network. 
The feedforward component of our network is a classic multilayer perceptron (MLP) with 3-layers, a binary activation function, and discrete (8 bit) weights. Our approach may, however, be extended to deeper networks and different encodings. 
In the following equations, each lowercase letter corresponds to a Boolean vector that represents a layer of spiking neurons on the chip (a spike corresponds to a 1).
The inference (forward) pass through the network is computed as:

\begin{align}
o & =  f(W_2 f(W_1 x)) \;,\label{eq:feedforward}  \\
f(x) & =  H(x-0.5)\;, \label{eq:binary_thr} \\
H(x) & =  \begin{cases} 0, & x < 0, \\ 1, & x \ge 0  \;, \end{cases} \label{eq:heaviside}
\end{align}
where $W_i$ is the weight matrix of the respective layer, $f$ is a binary activation function with a threshold of $0.5$, and $H$ denotes the Heaviside function. 
The forward pass thus occurs in 3 time steps as spikes are propagated through layers.
The degree to which the feedforward network’s output $(o)$ deviates from a target value $(t)$ is quantified by the squared error, $E = \frac{1}{2} \| o - t \|^2$, which we would like to minimize.  Performing backpropagation to achieve this requires the calculation of weight updates, which depend on the forward activations, and backward propagated local gradients $d_l$, which represent the amount by which the loss changes when the activity of that neuron changes, as:
\begin{eqnarray}
   d_2 & = & (o - t) \circ f'(W_2 h) \;,\\ \label{eq:error2}
   d_1 & = & \mathrm{sgn}(W_2^T d_2) \circ f'(W_1 x)\;,\\ \label{eq:error1}
   \frac{\partial{E}}{\partial{W_{l}}} & = & d_{l} (a_{l-1})^T \;,\\ \label{eq:grad}
   W_l^\mathrm{new} & = & W_l^\mathrm{old} - \eta \frac{\partial{E}}{\partial{W_{l}}} , \; l =1,2\;. \label{eq:updateEq}
\end{eqnarray} 
Here, $\circ$ denotes a Hadamard product, i.e. the element-wise product of two vectors, $^T$ denotes the matrix transpose, $\text{sgn}(x)$ is the sign function, and $a_l$ denotes the activation of the $l$th layer, $f(W_l a_{l-1})$, with $a_0 = x,\;a_1=h,\;a_2=o$. Here, $\eta$ denotes the learning rate, and is the only hyperparameter of the model apart from the weight initialization. 
Here $'$ denotes the derivative, but because $f$ is a binary thresholding function (Heaviside), the derivative would be the Dirac delta function, which is zero everywhere apart from at the threshold. Therefore, we use a common method \cite{bengio2013estimating, hubara2016binarized}, and represent the thresholding function using a truncated (between 0 and 1) ReLU (Eq.~\eqref{eq:trelu}) as a surrogate or straight-through estimator when back-propagating the error.  The derivative of the surrogate is a box function (Eq.~\eqref{eq:box}):
\begin{eqnarray}
f_\mathrm{surrogate}(x) & = & \min(\max(x, 0),1)\;, \label{eq:trelu}\\
f'(x) & = & H(x) - H(x-1)\;.\label{eq:box}
\end{eqnarray}
The three functions (Equations \eqref{eq:binary_thr}, \eqref{eq:trelu} and \eqref{eq:box}) are plotted in the inset in Fig.~\ref{fig:FuncConn}.

When performed for each target ($t$) in the training set, the model may be thought of as a stochastic gradient descent algorithm with a fixed step size update for each weight in the direction of the sign of the gradient. 

\subsubsection{sBP on Neuromorphic Hardware}
On the computational level, Equations \eqref{eq:feedforward}-\eqref{eq:box} fully describe our model exactly as it is implemented on Loihi, excluding the handling of bit precision constraints that affect integer discreteness and value limits and ranges.
In the following, we describe how these equations are translated from the computational to the algorithmic neural circuit level, thereby enabling implementation on neuromorphic hardware. Further details on the implementation can be found in Section~\ref{sec:methods}.

\begin{figure*}[!th]
  \centering
  \includegraphics[width=\linewidth]{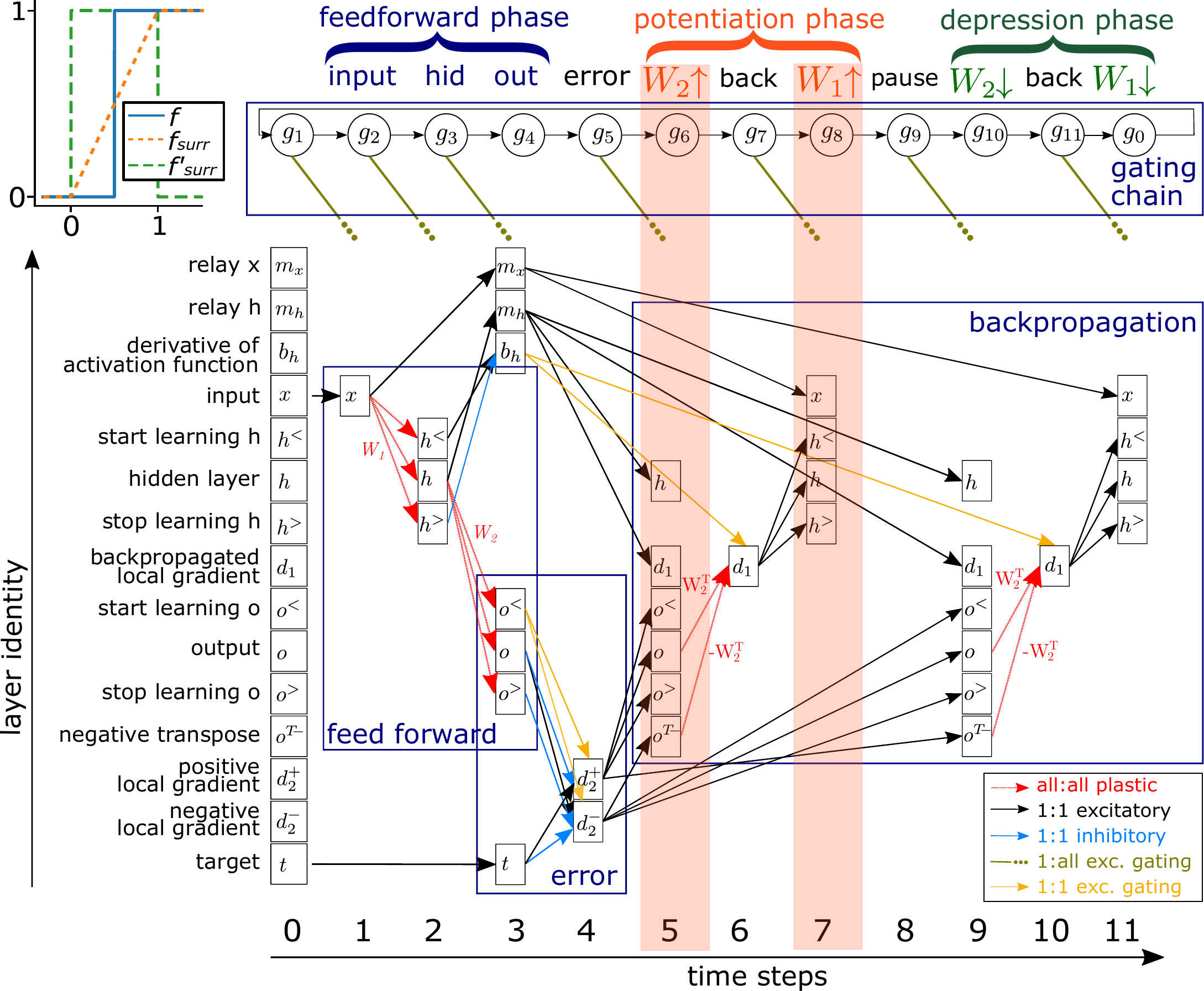}
  \caption{\textbf{Functional connectivity of the 2 layer backpropagation circuit.} Layers are only shown when they are gated `on' and synapses are only shown when their target is gated on.
  Plastic connections are all-to-all (fully connected), i.e. all neurons are connected to all neurons of the next layer. The gating connections from the gating chain are one-to-all, and all other connections are one-to-one, which means that a firing pattern is copied directly to the following layer. The names of the neuron layers are given on the left margin so that the row corresponds to layer identity. The columns correspond to time steps of the algorithm, which are the same as the time steps on Loihi. Table~\ref{tableMNISTcircuit} shows the information contained in each layer in each respective time step. The red background in time steps 5 and 7 indicates that in these steps, the sign of the weight update is inverted (positive), as $r=1$ in Eq.~\eqref{eq:learning_rule_loihi}. A detailed step-by-step explanation of the algorithm is given in Section~\ref{steps} and in Table~\ref{tableMNISTcircuit} in the supplementary material.
  The plot in the top left corner illustrates our approach to approximate the activation function $f$ by a surrogate with the box function as derivative, $f_{\text{surr}}=H(x)H(1-x)$, where $f'$ is the rectified linear map (ReLU) (see Equations~\ref{eq:binary_thr}, \ref{eq:trelu} and \ref{eq:box}).
  } \label{fig:FuncConn}
\end{figure*}

\paragraph{Hebbian weight update}
Equation \eqref{eq:updateEq} effectively results in the following weight update per single synapse from presynaptic index $i$ in layer $l-1$ to postsynaptic index $j$ in layer $l$:
\begin{equation}
\centering
 \label{eq:weight_update_single}
\begin{array}{c}
\Delta w_{ij} = -\eta \cdot a_{l-1,i} \cdot d_{l,j}\;,
 \end{array}
\end{equation}
where $\eta$ is the constant learning rate.
To accomplish this update, we use a Hebbian learning rule \cite{hebb_1949} implementable on the on-chip microcode learning engine (for the exact implementation on Loihi, see Section~\ref{sec:lr}).
Hebbian learning means that neurons that fire together, wire together, i.e., the weight update $\Delta w$ is proportional to the product of the simultaneous activity of the presynaptic (source) neuron and the postsynaptic (target) neuron. In our case, this means that the values of the 2 factors of Equation~\eqref{eq:weight_update_single} have to be propagated simultaneously, in the same time step, to the pre- ($a_{l-1,i}$) and postsynaptic ($d_{l,j}$) neurons while the pre- and postsynaptic neurons are not allowed to fire simultaneously at any other time.
For this purpose, a mechanism to control the information flow through the network is needed. 
\paragraph{Gating controls the information flow}
As our information control mechanism, we use synfire gating \cite{SornborgerTao1,SornborgerWangTao,WangSornborgerTao,wang2015fokker}. A closed chain of 12 neurons containing a single spike perpetually sent around the circle is the backbone of this flow control mechanism, which we call the gating chain.
The gating chain controls information flow through the controlled network by selectively boosting layers to bring their neurons closer to the threshold and thereby making them receptive to input. By connecting particular layers to the gating neuron that fires in the respective time steps, we lay out a path that the activity through the network is allowed to take. For example, to create the feedforward pass, the input layer $x$ is connected to the first gating neuron and therefore gated `on' in time step 1, the hidden layer $h$ is connected to the second gating neuron and gated `on' in time step 2, and the output layer $o$ is connected to the third gating neuron and gated `on' in time step 3. A schematic of this path of the activity can be found in Fig.~\ref{fig:FuncConn}. To speak in neuroscience terms, we are using synfire gating to design functional connectivity through the network anatomy shown in Supplementary Fig.~\ref{fig:AnaConn} .
Using synfire gating, the local gradient $d_{l,j}$ is brought to the postsynaptic neuron at the same time as the activity $a_{l-1,i}$ is brought back to the presynaptic neuron effecting a weight update.
However, in addition to bringing activity at the right time to the right place for Hebbian learning, the gating chain also makes it possible to calculate and back-propagate the local gradient.

\paragraph{Local gradient calculation }
For the local gradient calculation, according to Equation \eqref{eq:error2}, the error $o-t$ and the box function derivative of the surrogate activation function (Equation \eqref{eq:box}) are needed. 
Because there are no negative (signed) spikes, the local gradient is calculated and propagated back twice for a Hebbian weight update in two phases with different signs. The error $o-t$ is calculated in time step 4 in a layer that receives excitatory (positive) input from the output layer $o$ and inhibitory (negative) input from the target layer $t$, and vice versa for $t-o$.

The box function has the role of initiating learning when the presynaptic neuron receives a non-negative input and of terminating learning when the input exceeds 1, which is why we call the two conditions `start' and `stop' learning (inspired by the nomenclature of \cite{senn2005learning}). 
This inherent feature of backpropagation avoids weight updates that have no effect on the current output as the neuron is saturated by the nonlinearity with the current input. This regulates learning by protecting weights that are trained and used for inference when given different inputs.

To implement these two terms of the box function \eqref{eq:box}, we use two copies of the output layer that receive the same input ($W_2 h$) as the output layer. Using the above-described gating mechanism, one of the copies (start learning, $o^<$) is brought exactly to its firing threshold when it receives the input, which means that it fires for any activity greater than 0 and the input is not in the lower saturation region of the ReLU. The other copy (stop learning, $o^>$) is brought further away from the threshold (to 0), which means that if it fires, the upper saturation region of the ReLU has been reached and learning should cease.

\paragraph{Error backpropagation}
Once the local gradient $d_2$ is calculated as described in the previous paragraph, it is sent to the output layer as well as to its copies to bring about the weight update of $W_2$ and its 4 copies in time steps 5 and 9.
From there, the local gradient is propagated back through the transposed weight matrices $W_2^T$ and $-W_2^T$, which are copies of $W_2$ connected in the opposite direction and, in the case of $-W_2^T$, with opposite sign.
Once propagated backwards, the back-propagated error is also combined with the `start' and `stop' learning conditions, and then it is sent to the hidden layer and its copies in time steps 7 and 11.

\begin{figure}[!th]
  \centering
  \includegraphics[width=\linewidth]{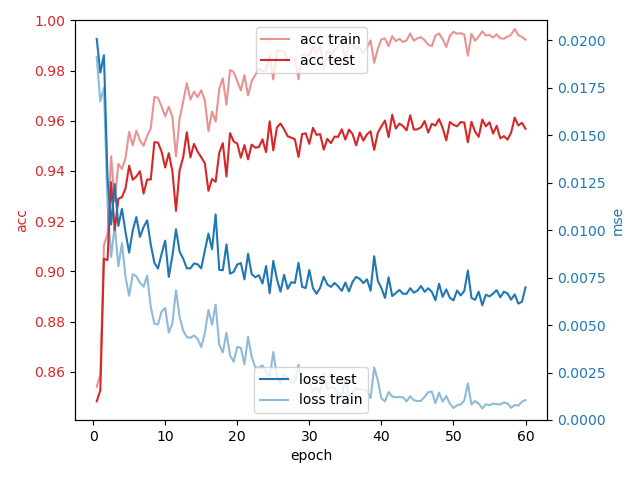}
  \caption{Accuracy and loss (mean squared error) over epochs. Note separate axis scaling for accuracy (left) and loss (right).}\label{fig:learning_curve}
\end{figure}

\subsubsection{Algorithm Performance}\label{sec:performance}
Our implementation of the sBP algorithm on Loihi achieves an inference accuracy of \finalaccuracy\% after \finalepoch\ epochs (best: \bestaccuracy\%) on the MNIST test data set, which is comparable with other shallow, stochastic gradient descent (SGD) trained MLP models without additional allowances. In these computational experiments, the sBP model is distributed over \LoihitrainingendnumCores\ neuromorphic cores. Processing of a single sample takes \Loihitrainingendlatencypertrial\ ms (\Loihiinferencelatencypertrial\ ms for inference only) on the neuromorphic cores, including the time required to send the input spikes from the embedded processor, and consumes \LoihitrainingendenergypertrialLoihi\ mJ of energy on the neuromorphic cores (\LoihitrainingendenergypertrialLoihiDynamic\ mJ of which is dynamic energy, i.e. energy used by our neural circuit in addition to the fixed background energy), resulting in an energy-delay product of \LoihitrainingendedpLoihiDynamic\ $\mu$Js. All measurements were obtained using NxSDK version 0.9.9 on the Nahuku32 board ncl-ghrd-01. Table~\ref{tab:lit_rev} shows a comparison of our results with published performance metrics for other neuromorphic learning architectures that were also tested on MNIST. Table~\ref{tab:energy} in the Supplementary Material shows a breakdown of energy consumption and a comparison of different conditions and against a GPU. Switching off the learning engine after training reduces the dynamic energy per inference to \LoihilearningoffenergypertrialLoihiDynamic\ mJ, which reveals that the on-chip learning engine is responsible for most of the power consumption.
Because the learning circuit is not necessary for performing inference, we also tested a reduced architecture that is able to do inference within four time steps using the previously trained weights. This architecture uses \LoihiinferenceenergypertrialLoihiDynamic\ mJ of dynamic energy and \Loihiinferencelatencypertrial\ ms per inference.

The sBP algorithm trains the network without explicit sparsity constraints, and yet it exhibits sparsity because of its binary (spiking) nature. After applying the binary threshold of 0.5 to the MNIST images, one image is encoded using on average 100 spikes in the input layer, which corresponds to a sparsity of 0.25 spikes per neuron per inference. This leads to a typical number of 110 spikes in the hidden layer (0.28 spikes per neuron per inference) and 1 spike in the output layer (0.1 spikes per neuron per inference). The spikes of the input and hidden layer are repeated in the two learning phases (see Fig. \ref{fig:FuncConn}) independent of the training progress. However, the error-induced activity from the local gradient layer $d_1$ starts with 0.7 spikes per neuron per sample (during the first 1000 samples) and goes down to approximately 0 spikes in the trained network as the error goes towards~0.

\begin{table*}
\footnotesize
\begin{tabular}{|l|l|l|l|l|l|l|}
\hline
Publication & Hardware & Learning Mode & Network & Energy per & Latency per& Test  \\
 &  &  &  Structure & Sample (mJ) & Sample (ms) & Accuracy (\%) \\
\hline
\multicolumn{7}{|l|}{\textbf{On-chip backpropagation}} \\ \hline
\textbf{This study} & Loihi & on-chip sBP & 400-400-10\footnote[1]{400 (20x20) corresponds to 784 (28x28) after cropping of the empty image margin of 4 pixels} & \LoihitrainingendenergypertrialLoihiDynamic & \Loihitrainingendlatencypertrial & \bestaccuracy \\
\hline
\multicolumn{7}{|l|}{\textbf{On-chip single layer training or BP alternatives}} \\ \hline
\cite{shrestha2021hardware} Shrestha et al. (2021)  &  Loihi & EMSTDP FA/DFA  & CNN-CNN-100-10 &  8.4  &  20 &   94.7 \\ 
\cite{frenkel202028} Frenkel et al. (2020) &  SPOON  & DRTP & CNN-10  & 0.000366\footnote[2]{Calculated from given values}  &  0.12 & 95.3 \\ 
\cite{park20197} Park et al. (2019) & unnamed & mod. SD & 784-200-200-10  & 0.000253\footnotemark[2] &  0.01 & 98.1 \\ 
\cite{chen20184096} Chen et al. (2018) &  unnamed & S-STDP &  236-20\footnote[3]{Off-chip preprocessing}  &    0.017     &  0.16 & 89   \\  
\cite{frenkel20180} Frenkel et al. (2018) &  ODIN  & SDSP & 256-10  & 0.000015 & - & 84.5 \\ 
\cite{lin2018programming} Lin et al. (2018) &  Loihi  & S-STDP& 1920-10\footnotemark[3]  & 0.553  & - & 96.4 \\ 
\cite{buhler20173} Buhler et al. (2017) & unnamed & LCA features & 256-10  & 0.000050  &  0.001\footnotemark[2] & 88 \\
\hline
\multicolumn{7}{|l|}{\textbf{On-chip inference only}} \\ \hline
\textbf{This study} & Loihi & inference & 400-400-10\footnotemark[1] & \LoihiinferenceenergypertrialLoihiDynamic & \Loihiinferencelatencypertrial   & \bestaccuracy \\
\cite{shrestha2021hardware} Shrestha et al. (2021) &  Loihi   & inference & CNN-CNN-100-10 &  2.47  &  10 &  94.7 \\
\cite{frenkel202028} Frenkel et al. (2020) &  SPOON  & inference & CNN-10  & 0.000313  &  0.12 & 97.5 \\ 
\cite{goltz2019fast} G\"oltz et al. (2019)  & BrainScaleS-2 & inference & 256-246-10   &    0.0084   &  0.048     &  96.9   \\
\cite{lin2018programming} Lin et al. (2018) &  Loihi  & inference & 1920-10\footnotemark[3]  & 0.0128\footnote[4]{Dynamic energy reported in the Supplementary Material of \cite{davies2021advancing}} & - & 96.4 \\ \cite{chen20184096} Chen et al. (2018)        &  unnamed &  inference &  784-1024-512-10  &    0.0017     & - & 97.9   \\ 
\cite{esser2015backpropagation} Esser et al. (2015)  & True North & inference &   CNN (512 neurons)  &  0.00027  &  1  &  92.7  \\
\cite{esser2015backpropagation} Esser et al. (2015) &   True North & inference & CNN (3840 neurons)    & 0.108   &  1   &   99.4  \\ 
\cite{stromatias2015scalable} Stromatias et al. (2015)  & SpiNNaker & inference  & 784-500-500-10  & 3.3  &  11   &   95  \\
\hline
\multicolumn{7}{|l|}{\textbf{Neuromorphic sBP in simulated SNN}} \\ \hline
\cite{jin2018hybrid} Jin et al. (2018)    & Simulation  & BP &   784-800-10   &  -  & -   & 98.8  \\
\cite{neftci2017event} Neftci et al. (2017)   & Simulation &  BP &  784-500-10 & - & - & 97.7 \\
\cite{shrestha2019approximating} Shrestha et al. (2019)        & Simulation &  EM-STDP  &   784-500-10 & -  &  -  &   97 \\
\cite{tavanaei2019bp} Tavanaei and Maida (2019)  &  Simulation  &  BP-STDP  & 784-500-150-10 &   -   &  -   &  97.2  \\
\cite{mostafa2017supervised} Mostafa (2017)   & Simulation & BP &  784-800-10   &  -  & -  & 97.55  \\
\cite{lee2016training} Lee et al. (2016)   & Simulation & BP &  784-800-10   &  -  & -  & 98.64  \\
\cite{oconnor2016deep} O'Connor and Welling (2016) & Simulation  & BP &  784-300-300-10   & -  & -  & 96.4 \\
\cite{diehl2015unsupervised} Diehl and Cook (2015)    &   Simulation  &  STDP  &     784-1600-10      &  -  & -   & 95 \\
\hline
\end{tabular} \label{literature_comparison}
\caption{Review of the MNIST Literature in SNN and on neuromorphic hardware. The table includes 4 relevant classes of literature.
Studies that used ANN-SNN conversion purely in software are not reviewed here. Note that the energy-delay product may be computed from the Energy per Sample and Latency per Sample columns.
Abbreviations:
EMSTDP: Error-modulated spike-timing dependent plasticity;
DFA: Direct feedback alignment;
DRTP: Direct random target projection;
SD: Segregated dendrites;
SDSP: Spike-driven synaptic plasticity;
LCA: Locally competitive algorithm.
}
\label{tab:lit_rev}
\end{table*}





\section{Discussion} 

\subsubsection{Summary} 

As we have demonstrated here, by using a well-defined set of neuronal and neural circuit mechanisms, it is possible to implement a spiking backpropagation algorithm on contemporary neuromorphic hardware. Previously proposed methods to address the issues outlined in Section~\ref{sec:main} were not on their own able to offer a straightforward path to implement a variant of the sBP algorithm on current hardware. In this study, we avoided or solved these previously encountered issues with spiking backpropagation by combining known solutions with synfire-gated synfire chains (SGSC) as a dynamical information coordination scheme that was evaluated on the MNIST test data set on the Loihi VLSI hardware.

\subsubsection{Solutions of Implementation Issues} The five issues (a)-(e) listed in Section~\ref{sec:main} were addressed using the following solutions:  (a) The \emph{weight transport} issue was avoided via the use of a deterministic, symmetric learning rule for the parts of the network that implement inference (feed-forward) and  error propagation (feedback) as described by \cite{ZipserRumelhart1990}. This approach is not biologically plausible because of a lack of developmental mechanisms to assure the equality of corresponding weights \cite{kolen1994backpropagation}. It would, however, without modifications to the architecture be feasible to employ weight decay as described by Kolen and Pollack \cite{kolen1994backpropagation} to achieve self-alignment of the backward weights to the forward weights or to use feedback alignment to approximately align the feedforward weights to random feedback weights \cite{lillicrap2016random}; (b) The \emph{backwards computation} issue was solved by using a separate error propagation network through which activation is routed using an SGSC; (c) The \emph{gradient storage} issue was solved by routing activity through the inference and error propagation circuits within the network in separate stages, thereby preventing the mixing of inference and error information. There are alternatives that would not require synfire gated routing, but are more challenging to implement on hardware \cite{payeur2020burst, sacramento2018dendritic} as also described in a more comprehensive review~\cite{lillicrap2020backpropagation}; (d) The \emph{differentiability} issue was solved by representing the step activation function by a surrogate in the form of a (truncated) ReLU activation function with an easily implementable box function derivative; and (e) The \emph{hardware constraint} issue was solved by the proposed mechanism's straightforward implementation on Loihi because it only requires basic integrate-and-fire neurons and Hebbian learning that is modulated by a single factor which is the same for all synapses. 

\subsubsection{Encoding and Sparsity}
While neural network algorithms on GPUs usually use operations on dense activation vectors and weight matrices, and therefore do not profit from sparsity, spiking neuromorphic hardware only performs an addition operation  when a spike event occurs, i.e., adding the weight to the input current as in Equation~\eqref{eq:loihi_eq}. This means that the power consumption directly depends on the number of spikes. Therefore sparsity, which refers to the property of a vector to have mostly zero elements, is important for neuromorphic algorithms \cite{davies2021advancing, stockl2021optimized}, and it is also observed in biology \cite{baddeley1997responses}. 
Consequently, significant effort has been made to make SNNs sparse to overcome the classical rate-based approach based on counting spikes \cite{stockl2021optimized, goltz2019fast, comsa2020temporal, rueckauer2018conversion}.
The binary encoding used here could be seen as a limit case of the rate-based approach allowing only 0 or 1 spike. Even without regularization to promote sparse activity, it yields very sparse activation vectors that are even sparser than most timing codes. The achievable encoded information per spike is unquestionably lower, however.
In a sense, we already employ spike timing to route spikes through the network because the location of a spike in time within the 12 time steps determines if and where it is sent and if the weights are potentiated or depressed.  However, usage of a timing code for activations could be enabled by having more than one Loihi time step per algorithm time step. Therefore the use of SGSCs is not limited to this particular binary encoding, and in fact, SGSCs were originally designed for a population rate code.

Similarly, the routing method we use in this work is not limited to backpropagation, but it could serve as a general method to route information in SNNs where autonomous activity (without interference from outside the chip) is needed. That is, our proposed architecture can act in a similar way as or even in combination with neural state machines \cite{Baumgartner2020Visual, liang2019neural}.

\subsubsection{Algorithmically-Informed Neurophysiology}
Although our implementation of sBP here was focused primarily on a particular hardware environment, we point out that the synfire-gated synfire chains and other network and neuronal structures that we employ could all potentially have relevance to the understanding of computation in neurophysiological systems. Many of these concepts that we use, such as spike coincidence, were originally inspired by neurophysiological experiments \cite{WangSornborgerTao,SornborgerWangTao,riehle1997spike}. Experimental studies have shown recurring sequences of stereotypical neuronal activation in several species and brain regions~\cite{abeles1993spatiotemporal,hahnloser2002ultra,ikegaya2004synfire} and particularly replay in hippocampus~\cite{foster2006reverse}. Recent studies also hypothesize~\cite{RajanHarveyTank2016, pang2019fast} and show \cite{malvache2016awake} that a mechanism like gating by synfire chains may play a role in memory formation. Additional evidence \cite{luczak2015packet} shows that large-scale cortical activity has a stereotypical, packet-like character that can convey information about the nature of a stimulus, or be ongoing or spontaneous. This type of apparently algorithmically-related activity has a very similar form to the SGSC controlled information flow found previously \cite{SornborgerWangTao,WangSornborgerTao,wang2015fokker,XiaoWangSornborgerTao}.  Additionally, this type of sequential activation of populations is evoked by the sBP learning architecture, as seen in the raster plot in Fig.~\ref{fig:Raster} in the Supplementary Material.

Other algorithmic spiking features, such as the back-propagated local gradient layer activity decreasing from 0.7 spikes per neuron to 0 by the end of training, could be identified and used to generate qualitative and quantitative hypotheses concerning network activity in biological neural systems.

\subsubsection{Future Directions}

Although the accuracy we achieve is similar to early implementations of binary neural networks in software \cite{hubara2016binarized},  subsequent approaches now reach ~98.5\% \cite{simons2019review}, and generally include binarized weights. However, networks that achieve such accuracy typically employ either a convolutional structure or multiple larger hidden layers. Additional features such as dropout, softmax final layers, gain terms, and others could in principle be included in spiking hardware and may also account for this 3\% gap.
So, while we show that it is possible to efficiently implement backpropagation on neuromorphic hardware, several non-trivial steps are still required to make it usable in practical applications: 

1) The algorithm needs to be scaled to deeper networks. While the present structure is in principle  scalable to more layers without major adjustments, investigation is needed to determine whether gradient information remains intact over many layers, and to what extent additional features such as alternatives to batch normalization may need to be developed.

2) Generalization to convolutional networks is compelling, in particular for application to image processing.  The Loihi hardware presents an advantage in this setting because of its weight-sharing mechanisms.

3) Although our current implementation demonstrates on-chip learning, we train on the MNIST images in an offline fashion by iterating over the training set in epochs.  Further research is required to develop truly continual learning mechanisms such that additional samples and classes can actually be learned without losing previously trained synaptic weights and without retraining on the whole dataset.

Additionally, the proposed algorithmic methodology can be used to inform hardware adjustments to promote efficiency for learning applications.  Although our algorithm is highly efficient in terms of power usage, in particular for binary encoding, the Loihi hardware is not specifically designed for implementing standard deep learning models, but rather as general-purpose hardware for exploring different SNN applications \cite{davies2021advancing}.

This leads to a significant computational overhead for functions that are not needed in our model (e.g. neuronal dynamics), or that could have been realized more efficiently if integrated directly on the chip instead of using network mechanisms. Our results provide a potential framework to guide future hardware modifications to facilitate more efficient learning algorithm implementations. For example, in an upgraded version of the algorithm, it would be preferable to replace relay neurons with presynaptic (eligibility) traces to keep a memory of the presynaptic activity for the weight update.

\subsubsection{Significance}\label{sec:significance}
To our knowledge, this work is the first to show an SNN implementation of the backpropagation algorithm that is fully on-chip, without a computer in the loop. Other on-chip learning approaches so far either use feedback alignment \cite{shrestha2021hardware}, forward propagation of errors \cite{frenkel202028} or single layer training \cite{stewart2020chip,frenkel20180,lin2018programming,chen20184096,buhler20173}.
Compared to an equivalent implementation on a GPU, there is no loss in accuracy, but there are about two orders of magnitude power savings in the case of small batch sizes which are more realistic for edge computing settings.
So, this implementation shows a path for using in-memory, massively parallel neuromorphic processors for low-power, low-latency implementation of modern deep learning applications.
The network model we propose offers opportunities as a building block that can, e.g. be integrated into larger SNN architectures that could profit from a trainable on-chip processing stage.

\section{Methods} \label{sec:methods}
In this section, we describe our system on three different levels~\cite{marr1976understanding}.
First, we describe the computational level by fully stating the equations that result in the intended computation. Second, we describe the spiking neural network (SNN) algorithm.
Third, we describe the details of our hardware implementation that are necessary for exact reproducibility.

\subsection{The Binarized Backpropagation Model}\label{sec:model}
\noindent
\textbf{Network Model.} Backpropagation is a means of optimizing synaptic weights in a multi-layer neural network. It solves the problem of credit assignment, i.e., attributing changes in the error to changes in upstream weights, by recursively using the chain rule. 
The inference (forward) pass through the network is computed as
\begin{align}\label{netout}
o = f(W_N f(W_{N-1}(\dots f(W_1 x))))\;,
\end{align}
where $f$ is an element-wise nonlinearity and $W_i$ is the weight matrix of the respective layer. 
The degree to which the network’s output ($o$) deviates from the target values ($t$) is quantified by the squared error, $E = \frac{1}{2} \| o - t \|^2$, which we aim to minimize. The weight updates for each layer are computed recursively by
\begin{eqnarray}
  d_l & = & \left\{ \begin{array}{ll} (o - t) \circ  f'(n_l), & l = N \\ 
            W_{l+1}^T d_{l+1} \circ  f'(n_l), & l < N \end{array} \right.\;, \label{eq:deltajay}\\
  \frac{\partial{E}}{\partial{W_{l+1}}} & = & d_{l+1} (a_l)^T, \\
  W_l^\mathrm{new} & = & W_l^\mathrm{old} - \eta \frac{\partial{E}}{\partial{W_{l}}} \;, \label{eq:updateEq2}
\end{eqnarray} 
where $n_l$ is the network activity at layer $l$ ({\it i.e.,} $n_l = W_l f(W_{l-1} n_{l-1})$). Here, $'$ denotes the derivative, $\circ$ denotes a Hadamard product, i.e. the element-wise product of two vectors, $^T$ denotes the matrix transpose, and $a_l$ denotes $f(W_l a_{l-1})$, with $a_0 = x$. The parameter $\eta$ denotes the learning rate. These general equations \eqref{eq:deltajay}-\eqref{eq:updateEq2} are realized for two layers in our implementation as given by \eqref{eq:error2}-\eqref{eq:updateEq} in Section~\ref{sec:results}.  Below, we relate these equations to the neural Hebbian learning mechanism used in the neuromorphic implementation of sBP. 

Although in theory the derivative $f'$ of the activation function is applied in \eqref{eq:deltajay},  in the case that $f$ is a binary thresholding (Heaviside) function, the derivative is the Dirac delta function, which is zero everywhere apart from at the threshold. We use a common approach \cite{hubara2016binarized} and represent the activation function by a surrogate (or straight-through estimator~\cite{bengio2013estimating}), in the form of a rectified linear map (ReLU) truncated between 0 and 1 (Eq.~\eqref{eq:trelu2}) in the part of the circuit that affects  error backpropagation. The derivative of this surrogate ($f'$) is of box function form, i.e.
\begin{eqnarray}
f(x) & = & H(x-0.5) \label{eq:binary_thr2}\;, \\
f_\mathrm{surrogate}(x) & = & \min(\max(x, 0),1)\;, \label{eq:trelu2}\\
f'(x) & = & H(x) - H(x-1)\;, \label{eq:box2}
\end{eqnarray}
where $H$ denotes the Heaviside function:
\begin{eqnarray}
H(x)=\begin{cases} 0, & x < 0 \;, \\ 1, & x \ge 0 \;. \end{cases}
\end{eqnarray}
In the following section, we describe how we implement Equations \eqref{netout}-\eqref{eq:trelu2} in a spiking neural network.

\noindent
\textbf{Weight Initialization.} \label{weight_init}
Plastic weights are initialized by sampling from a Gaussian distribution with mean of 0 and a standard deviation of $1/\sqrt{2/(N_\mathrm{fanin}+N_\mathrm{fanout})}$ (He initialization \cite{he2015delving}). $N_\mathrm{fanin}$ denotes the number of neurons of the presynaptic layer and $N_\mathrm{fanout}$ the number of neurons of the postsynaptic layer.

\noindent
\textbf{Input Data.}
The images of the MNIST dataset \cite{lecun1998gradient} were cropped by a margin of 4 pixels on each side to remove pixels that are never active and avoid unused neurons and synapses on the chip. The pixel values were thresholded with 0.5 to get a black and white picture for use as input to the network. In the case of the $100-300-10$ architecture, the input images were downsampled by a factor of 2.
The dataset was presented in a different random order in each epoch.

\noindent
\textbf{Accuracy Calculation.}
The reported accuracies are calculated on the full MNIST test data set. A sample was counted as correct when the index of the spiking neuron of the output layer in the output phase (time step 3 in Fig.~\ref{fig:FuncConn}) is equal to the correct label index of the presented image. In fewer than 1\% of cases, there was more than one spike in the output layer, and in that case, the lowest spiking neuron index was compared.

\subsection{The Spiking Backpropagation Algorithm}
\label{sec:spiking_algo}

\noindent
\textbf{Spiking Neuron Model.} \label{neuron_model}
For a generic spiking neuron element, we use the current-based linear leaky integrate-and-fire model (CUBA). This model is implemented on Intel's neuromorphic research processor, codenamed Loihi \cite{DBLP:journals/micro/DaviesSLCCCDJIJ18}.
Time evolution in the CUBA model as implemented on Loihi is described by the discrete-time dynamics with $t\in\mathds{N}$, and time increment $\Delta t \equiv 1$:
\begin{equation}
\label{eq:loihi_eq}
\begin{array}{cc}
V_{i}(t+1) &= V_{i}(t) -\frac{1}{\tau_V} V_{i}(t) + U_{i}(t) + I_{\mathrm{const}}\;,
\\[8pt]
U_{i}(t+1) &= U_{i}(t)-\frac{1}{\tau_U} U_{i}(t) + \sum_{j} w_{ij} \delta_{j}(t) \; ,
 \end{array}
\end{equation}
where $i$ identifies the neuron.

The membrane potential $V(t)$ is reset to $0$ upon exceeding the threshold $V_\mathrm{thr}$ and remains at its reset value $0$ for a refractory period, $T_\mathrm{ref}$. Upon reset, a spike is sent to all connecting synapses. Here, $U(t)$ represents a neuronal current and $\delta$ represents time-dependent spiking input. $I_{\mathrm{const}}$ is a constant input current.

\noindent
\textbf{Parameters and Mapping.}
In our implementation of the backpropagation algorithm, we take $\tau_V = \tau_U = 1$, $T_\mathrm{ref} = 0$, and $I_\mathrm{const} = -8192$ (except in gating neurons, where $I_\mathrm{const} = 0$).
This leads to a memoryless point neuron that spikes whenever its input in the respective time step exceeds $V_\mathrm{thr} = 1024$. This happens, when the neuron receives synaptic input larger than $0.5 \cdot V_\mathrm{thr}$ and in the same timestep, a gating input overcomes the strong inhibition of the $I_\mathrm{const}$, i.e. it is gated `on'. This is how the Heaviside function in Equation~\eqref{eq:binary_thr2} is implemented. For the other activation functions, a different gating input is applied.

There is a straightforward mapping between the weights and activations in the spiking neural network (SNN) described in this section and the corresponding artificial neural network (ANN) described in Section~\ref{sec:model}:

\begin{equation}
\label{eq:weight_mapping}
\begin{array}{cc}
w_\mathrm{SNN} = w_\mathrm{ANN} \cdot V_\mathrm{thr}\;, \\
a_\mathrm{SNN} = a_\mathrm{ANN} \cdot V_\mathrm{thr}\;.
\end{array}
\end{equation}
So, a value of $V_\mathrm{thr}= 1024$ allows for a maximal ANN weight of 0.25, because the allowed weights on Loihi are the even integers from -256 to 254.

\noindent
\textbf{Feed-forward Pass.}
The feedforward pass can be seen as an independent circuit module that consists of 3 layers. An input layer $x$ with 400 (20x20) neurons that spikes according to the binarized MNIST dataset, a hidden layer $h$ of 400 neurons, and an output layer $o$ of 10 neurons. The 3 layers are sequentially gated `on' by the gating chain so that activity travels from the input layer to the hidden layer through the plastic weight matrix $W_1$ and then from the hidden to the output layer through the plastic weight matrix $W_2$.

\noindent
\textbf{Learning Rule.} \label{sec:lr}
The plastic weights follow the standard Hebbian learning rule with a global third factor to control the sign.
Note, however, that here, unlike other work with Hebbian learning rules, due to the particular activity routed to the layers, the learning rule implements a supervised mechanism (backpropagation).
Here we give the discrete update equation as implemented on Loihi:
\begin{eqnarray}
\Delta w & = & 4 r(t) \circ x(t) \circ y(t) - 2 x(t) \circ y(t) \label{eq:learning_rule_loihi}\\ 
  & = & (2r(t)-1) \cdot 2 x(t) y(t) \;, \nonumber \\
\mathrm{r}(t) & = & \begin{cases}
    1 ,& \text{if  } (t\bmod T) =  {5,7}\;,\\
0 ,& \text{ otherwise} \;.
\end{cases}
\end{eqnarray}
Above, $x$, $y$, and $r$ represent time series that are available at the synapse on the chip. The signals $x$ and $y$ are the pre- and postsynaptic neuron's spike trains, i.e., they are equal to 1 in time steps when the respective neuron fires, and 0 otherwise. The signal $r$ is a third factor that is provided to all synapses globally and determines in which phase (potentiation or depression) the algorithm is in. 
$T$ denotes the number of phases per trial, which is 12 in this case. So, $r$ is 0 in all time steps apart from the \nth{5} and \nth{7} of each iteration, where the potentiation of the weights happens. This regularly occurring $r$ signal could thus be generated autonomously using the gating chain. On Loihi, $r$ is provided using a so-called ``reinforcement channel".
Note that the reinforcement channel can only provide a global modulatory signal that is the same for all synapses.

The above learning rule produces a positive weight update in time steps in which all three factors are 1, i.e., when both pre- and post-synaptic neurons fire and the reinforcement channel is active. It produces a negative update when only the pre- and post-synaptic neurons fire, and the weight stays the same in all other cases.

To achieve the correct weight update according to the backpropagation algorithm (see \eqref{eq:weight_update_single}), the spiking network has to be designed in a way that the presynaptic activity $a_{l-1,i}$ and the local gradient $d_{l,j}$ are present in neighboring neurons at the same time step. Furthermore, the sign of the local gradient has to determine if the simultaneous activity happens in a time step with active third factor $r$ or not.

This requires control over the flow of spikes in the network, which is achieved by a mechanism called synfire gating \cite{ShaoSornborgerTao,ShaoWangSornborgerTao2018}, which we adapt and simplify here.

\noindent
\textbf{Gating Chain.} Gating neurons are a separate structure within the backpropagation circuit and are connected in a ring. That is, the gating chain is a circular chain of neurons that, once activated, controls the timing of information processing in the rest of the network. This allows information routing throughout the network to be autonomous to realize the benefits of neuromorphic hardware without the need for control by a classical sequential processor. Specifically, the neurons in the gating chain are connected to the relevant layers of the network, which allows them to control when and where information is propagated. All layers are inhibited far away from their firing threshold by default, as described in Section~\ref{neuron_model}, and can only transmit information, i.e., generate spikes, if their inhibition is neutralized via activation by the gating chain.  Because a neuron only fires if it is gated `on' AND gets sufficient input, such gating corresponds to a logical AND or coincidence detection with the input.

In our implementation, the gating chain consists of 12 neurons, which induce 12 algorithmic (Loihi) time steps that are needed to process one sample. Each neuron is connected to all layers that must be active in each respective time step. 
The network layers are connected in the manner shown in Supplementary Fig.~\ref{fig:AnaConn}, but the circuit structure, which is controlled by the gating chain, results in a functionally connected network as presented in Fig.~\ref{fig:FuncConn}, where the layers are shown according to the timing of when they become active during one iteration.

The weight of the standard gating synapses (from one of the gating neurons to each neuron in a target layer) is $w_\mathrm{gate} = -I_\mathrm{const} + 0.5 V_\mathrm{thr}$, i.e. each neuron that is gated `on' is brought to half of its firing threshold, which effectively implements Eq.~\eqref{eq:binary_thr2}.
In four cases, i.e., for the synapses to the start learning layers in time step 2 ($h^<$) and 3 ($o^<$) and to the backpropagated local gradient layer $d_1$ in time steps 6 and 10, the gating weight is $w_\mathrm{gate} = -I_\mathrm{const} + V_\mathrm{thr}$. In two cases, i.e., for the synapses to the stop learning layer in time step 2 ($h^>$) and 3 ($o^>$), the gating weight is $w_\mathrm{gate} = -I_\mathrm{const}$.
These different gating inputs lead to step activation functions with different thresholds, as required for the computations explained below, in Section \ref{start_stop}.

\noindent
\textbf{Backpropagation Network Modules.}
In the previous sections, we have explained how the weight update happens and how to bring the relevant values ($a_{l-1}$ and $d_{l}$ according to  \eqref{eq:weight_update_single}) to the correct place at the correct time. In this section, we discuss how these values are actually calculated. The signal $a_{l-1}$, which is the layer activity from the feedforward pass, does not need to be calculated but only remembered. This is done using a relay layer with synaptic delays, as explained in Section~\ref{relay}.  The signal $d_{2}$, the last layer local gradient, consists of 2 parts according to \eqref{eq:error2}. The difference between the output and the target $o-t$ (see Section~\ref{error_calc}) and the box function $f'$ must be calculated. We factorize the latter into two terms, a start and stop learning signal (see Section~\ref{start_stop}).
The signal $d_{1}$, the backpropagated local gradient, also consists of 2 parts, according to Eq.~\eqref{eq:error1}. In addition to another `start' and `stop' learning signal, we need $\mathrm{sgn}(W_2^T d_2)$, whose computation is explained in Section~\ref{backprop}.

In the following equation, the weight update is annotated with the number of the paragraph in which the calculating module is described:

\begin{eqnarray}
\Delta W_2 &=& \eta \overbrace{(o-t)}^{\text{\hyperref[error_calc]{2.}}} \circ \overbrace{f'(W_2h)}^{\text{\hyperref[start_stop]{3.}}}  \overbrace{h^T}^{\text{\hyperref[relay]{1.}}}\;,\\
\Delta W_1 &=& \eta \underbrace{\mathrm{sgn}(W_2^T d_2)}_{\text{\hyperref[backprop]{4.}}} \circ \underbrace{f'(W_1 x)}_{\text{\hyperref[start_stop]{3.}}} \underbrace{x^T}_{\text{\hyperref[relay]{1.}}} \;.
\end{eqnarray}

\subsubsection{Relay Neurons} \label{relay}
The memory used to store the activity of the input and the hidden layer is a single relay layer that is connected both from and to the respective layer in a one-to-one manner with the proper delays. The input layer $x$ sends its activity to the relay layer $m_x$ so that the activity can be restored in the $W_1$ update phases in time steps 7 and 11. The hidden layer $x$ sends its activity to the relay layer $m_h$ so that the activity can be restored in the $W_2$ update phases in time steps 5 and 9.

\subsubsection{Error Calculation} \label{error_calc}
The error calculation requires a representation of signed quantities, which is not directly possible in a spiking network because there are no negative spikes. This is achieved here by splitting the error evaluation into two parts, $t-o$ and $o-t$, to yield the positive and negative components separately. Similarly, the calculation of back-propagated local gradients, $d_{1}$, is performed using a negative copy of the transpose weight matrix, and it is done in 2 phases for the positive and negative local gradient, respectively.
In the spiking neural network, $t-o$ is implemented by an excitatory synapse from $t$ and an inhibitory synapse of the same strength from $o$, and vice versa for $o-t$. Like almost all other nonplastic synapses in the network, the absolute weight of the synapses is just above the value that makes the target neuron fire, when gated on.
The difference between the output and the target is, however, just one part of the local gradient $d_2$. The other part is the derivative of the activation function (box function).

\subsubsection{Start and Stop Learning Conditions} \label{start_stop}
The box function \eqref{eq:box2} can be split in two conditions: a `start' learning and a `stop' learning condition. These two conditions are calculated in parallel with the feedforward pass. The feedforward activation $f(x)=H(x-0.5 V_\mathrm{thr})$ corresponding to  Eq.~\eqref{eq:feedforward} is an application of the spiking threshold to the layer's input with an offset of $0.5 V_\mathrm{thr}$, which is given by the additional input from the gating neurons.
The first term of the box function \eqref{eq:box}, $H(x)$, is also an application of the spiking threshold, but this time with an offset equal to the firing threshold so that any input larger than 0 elicits a spike. We call this first term the `start' learning condition, and it is represented in $h^<$ for the hidden and in $o^<$ for the output layer.
The second term of the box function in Eq.~\eqref{eq:box}, $-H(x-1 V_\mathrm{thr})$, is also an application of the spiking threshold, but this time without an offset so that only an input larger than the firing threshold elicits a spike. We call this second term the `stop' learning condition, and it is represented in $h^>$ and $o^>$ for the hidden and output layers, respectively.
For the $W_1$ update, the two conditions are combined in a box function layer $b_h= h^{<} - h^{>}$ that then gates the $d_1$ local gradient layer.
For the $W_2$ update, the two conditions are directly applied to the $d_2$ layers because an intermediate $b_o$ layer would waste one time step. The function is however the same: the stop learning $o^>$ inhibits the $d_2$ layers and the `start' learning signal $o^<$ gates them.  
In our implementation, the two conditions are obtained in parallel with the feedforward pass, which requires two additional copies of each of the two weight matrices. An alternative method to avoid these copies but takes more time steps, would do this computation sequentially and use the synapses and layers of the feedforward pass three times per layer with different offsets, and then route the outputs to their respective destinations.

\subsubsection{Error Backpropagation} \label{backprop}
Error calculation and gating by the start learning signal and inhibition by the stop learning signal are combined in time step 4 to calculate the last layer local gradients $d^+_2$ and $d^-_2$. From there, the local gradients are routed into the output layer and its copies for the last layer weight update. This happens in 2 phases: The positive local gradient $d^+_2$ is connected without delay so that it leads to potentiation of the forward and backward last layer weight matrices in time step 5. The negative local gradient is connected with a delay of 4 time steps so that it arrives in the depression phase in time step 9. For the connections to the $o^{T-}$ layer which is connected to the negative copy $-W_2^T$, the opposite delays are used to get a weight update with the opposite sign. See Fig.~\ref{fig:FuncConn} for a visualization of this mechanism.
Effectively, this leads to the last layer weight update

\begin{align}
\label{eq:2phaseupdate_w2}
\Delta W_{2} = \eta (H(t-o) \circ f'(W_2 h)) h^T \nonumber \\
- \eta (H(o-t) \circ f'(W_2 h)) h^T\;,
\end{align}
where the first term on the right hand side is non-zero when the local gradient is positive, corresponding to an update happening in the potentiation phase, and the second term is nonzero when the local gradient is negative, corresponding to an update happening in the depression phase. The functions $f$ and $f'$ are as described in Equations \eqref{eq:binary_thr2} and \eqref{eq:box}.

The local gradient activation in the output layer does not only serve the purpose of updating the last layer weights, but it is also directly used to backpropagate the local gradients.
Propagating the signed local gradients backwards through the network layers requires a positive and negative copy of the transposed weights, $W_2^{T}$ and $-W_2^{T}$, which are the weight matrices of the synapses between the output layer $o$ and the back-propagated local gradient layer $d_1$, and between $o^{T-}$ and $d_1$, respectively. Here $o^{T-}$ is an intermediate layer that is created exclusively for this purpose.
The local gradient is propagated backwards in two phases. The part of the local gradient that leads to potentiation is propagated back in time steps 5 to 7, and the part of the local gradient that leads to depression of the $W_1$ weights is propagated back in time steps 9 to 11.
In time step 6, the potentiating local gradient is calculated in layer $d_1$ as
\begin{equation}
    d_1^+ = H(W_2^{T}d_2^+ + (-W_2^{T})d_2^-)\circ b_h\;,
\end{equation}
and in timestep 10 the depressing local gradient is calculated in layer $d_1$ as
\begin{equation}
    d_1^- = H((-W_2^{T})d_2^+ + W_2^{T}d_2^-)\circ b_h\;.
\end{equation}
Critically, this procedure does \emph{not} simply separate the weights by sign, but rather maintains a complete copy of the weights that is used to associate appropriate sign information to the back-propagated local gradient values. Note that here the Heaviside activation function $H(x)$ is used rather than the binary activation function $f=H(x-0.5 V_\mathrm{thr})$, so that \emph{any} positive gradient will induce an update of the weights. Any positive threshold in this activation will lead to poor performance by making the learning insensitive to small gradient values. 
The transposed weight copies must be kept in sync with their forward counterparts, so the updates in the potentiation and depression phases are also applied to the forward and backward weights concurrently.

So in total, after each trial, the actual first layer weight update is the sum of four different parts:
\begin{align}
\centering
\label{eq:2phaseupdate_w1}
\Delta W_{1} & =  \eta ((H(W_2^{T} d_2^+ + (-W_2^{T}) d_2^-))\circ b_h)  x^T\nonumber \\ & \quad - \eta ((H((-W_2^{T}) d_2^+ + W_2^{T} d_2^-))\circ b_h) x^T.
\end{align}

These four terms are necessary because, e.g., a positive error can also lead to depression if backpropagated through a negative weight matrix and the other way round.

\subsection{The sBP Implementation on Loihi}

\noindent
\textbf{Partitioning on the Chip.}
\label{partitioning}
To distribute the spike load over cores, neurons of each layer are approximately equally distributed over 96 cores of a single chip. This distribution is advantageous because only a few layers are active at each time step, and Loihi processes spikes within a core in a sequential manner.
In total, the network as presented here needs 2$N_\mathrm{in}$+6$N_\mathrm{hid}$+7$N_\mathrm{out}$+12$N_\mathrm{gat}$ neurons. With $N_\mathrm{in} = 400$, $N_\mathrm{hid}=400$, $N_\mathrm{out}=10$, and $N_\mathrm{gat}=1$, these are 3282 neurons and about 200k synapses, most of which are synapses of the 3 plastic all-to-all connections between the input and the hidden layer. 

\noindent
\textbf{Learning Implementation.}
The learning rule on Loihi is implemented as given in Eq.~\eqref{eq:learning_rule_loihi}.
Because the precision of the plastic weights on Loihi is maximally 8 bits with a weight range from $-256$ to $254$, we can only change the weight in steps of 2 without making the update non-deterministic. This is necessary for keeping the various copies of the weights in sync (hence the factor of 2 in Eq.~\eqref{eq:learning_rule_loihi}). With $V_\mathrm{thr} = 1024$, this corresponds to a learning rate of $\eta = \frac{2}{1024} \approx 0.002$. The learning rate can be changed by increasing the factor in the learning rule, which leads to a reduction in usable weight precision, or by changing $V_\mathrm{thr}$, which changes the range of possible weights according to Eq. \eqref{eq:weight_mapping}.
Several learning rates (settings of $V_\mathrm{thr}$) were tested with the result that the final accuracy is not very sensitive to small changes. The learning rate that yielded the best accuracy is reported here. 
In the NxSDK API, the neuron traces that are used for the learning rule are \texttt{x0}, \texttt{y0}, and  \texttt{r1} for $x(t)$, $y(t)$ and $r(t)$ in \ref{eq:learning_rule_loihi} respectively. \texttt{r1} was used with a decay time constant of 1, so that it is only active in the respective time step, effectively corresponding to \texttt{r0}.
To provide the $r$ signal, a single reinforcement channel was used and was activated by a regularly firing spike generator in time steps 5 and 7.

\noindent
\textbf{Weight Initialization.}
After the He initialization as described in Section~\ref{weight_init}, the weights are mapped to Loihi weights according to \eqref{eq:weight_mapping}. Then, the weights are truncated to $[-240, 240]$ and discretized to 8 bits resolution, i.e, steps of 2, by rounding them to the next valid number towards 0.

\noindent
\textbf{Power Measurements.}
All Loihi power measurements  are  obtained  using  NxSDK version 0.9.9 on the Nahuku32 board ncl-ghrd-01. Both software API and hardware were provided by Intel Labs. All other probes, including the output probes, are deactivated. For the inference measurements, we use a network that only consists of the three feedforward layers with non-plastic weights and the gating chain of four neurons. The power was measured for the first 10000 samples of the training set for the training measurements and all 10000 samples of the test set for the inference measurements.

\bibliography{Biblio,neuromorphicBP}

\section*{Acknowledgements}
This work was carried out at Los Alamos National Laboratory under the auspices of the National Nuclear Security Administration of the U.S. Department of Energy under Contract No. 89233218CNA000001 - SEEK: Scoping neuromorphic architecture impact enabling advanced sensing capabilities. Additional funding was provided by the LANL ASC Beyond Moore's Law program and by LDRD Reserve Project 20180737ER  - Neuromorphic Implementation of the Backpropagation Algorithm.  L.T. acknowledges support from the Natural Science Foundation of China grants 31771147 (L.T.) and 91232715 (L.T). L.T. thanks the Los Alamos National Laboratory for its hospitality. 
A.R. acknowledges support from the Swiss National Science Foundation (SNSF) grant Ambizione PZOOP2 168183 ELMA. F.S. carried out work under the kind support of the Center for Nonlinear Studies fellowship and final preparations at the London Institute for Mathematical Sciences. The Authors thank Intel Labs for providing support and access to the Loihi API and hardware. 

\section*{Author Contributions}
The authors contributed equally to the methodology presented here.  Alpha Renner adapted and implemented the algorithm on Intel's Loihi chip. Alpha Renner, Forrest Sheldon, and Anatoly Zlotnik formalized neuromorphic information-processing mechanisms, implemented the algorithm in simulation and hardware, and developed figures. All authors wrote the manuscript with Renner, Sheldon, and Zlotnik doing the bulk of the writing.  Andrew Sornborger and Anatoly Zlotnik supervised the research and Sornborger and Louis Tao developed the concepts and algorithmic basis of the neuromorphic backpropagation algorithm and circuit structure.  

\section*{Competing Financial Interests}
The authors declare that they have no competing financial interests.

\clearpage
\newpage

\begin{widetext}
\ifdefined\linenumbers \begin{nolinenumbers} \fi
\section*{Supplementary Material} \label{sec:supp}

\begin{figure*}[h]
  \centering
  \includegraphics[width=0.6\columnwidth]{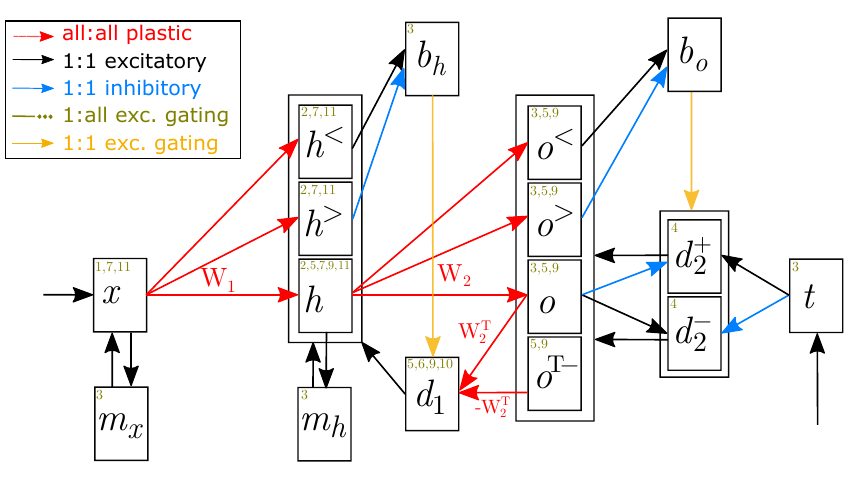}
  \caption{\textbf{Anatomical connectivity of the 2 layer backpropagation circuit.} 
  While in the Loihi implementation the $o$ layers are connected directly go to the $d_2$ layers, here an intermediate fictional $b_o$ layer is added for easier understanding. Arrows that end on the border of a box that encompasses several layers go to each of the layers. The gating chain is not shown, but the small numbers on top of each layer indicate when it is gated on. Colors are the same as in Fig.~\ref{fig:FuncConn}.
  } \label{fig:AnaConn}
\end{figure*}

\clearpage

\begin{figure*}[!th]
  \centering
  \includegraphics[width=\columnwidth]{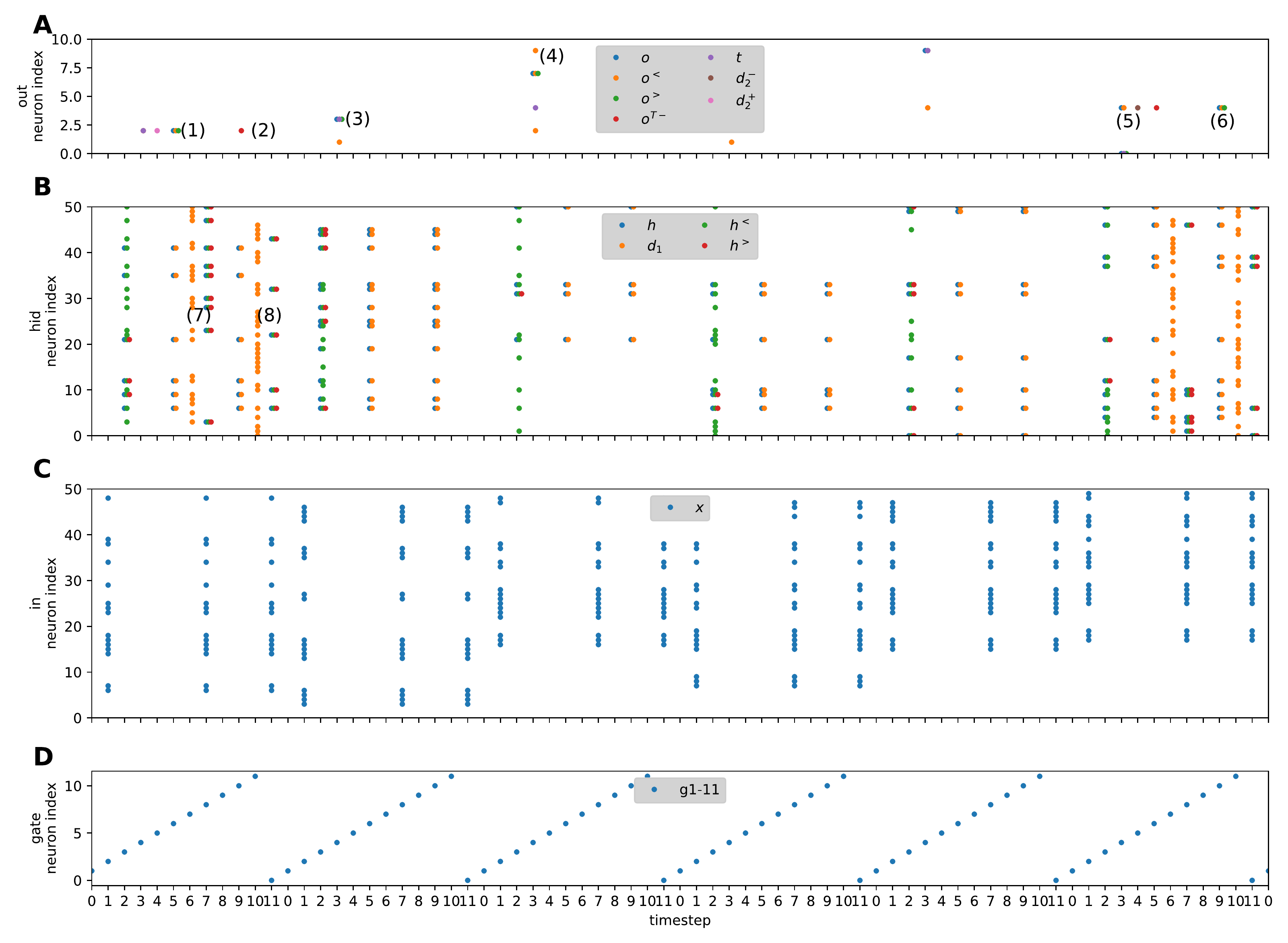}
  \caption{Example raster plot of the spikes over six gating cycles. All populations of the same size are plotted in the same plot and only the first 50 neurons are plotted per layer. To avoid occlusion, a small offset in time is added to the time step of some layers. Refer to Tab.~\ref{tableMNISTcircuit} for a detailed explanation of the spike propagation. (1) error (target but no output spike) leads to potentiation of the $W_2$ synaptic weight and the positive transpose; (2) the same error leads to depression of the negative transpose via activity of $d_1$; (3) no error because $o$ and $t$ fire at the same location, i.e. there is no update in this iteration; (4) there is an error ($t$ fires at index 4, but $o$ at index 7), but the local gradient is 0 because it is gated `off' at index 7 because the derivative of the activation function is 0, i.e. both $o^<$ and $o^>$ fire. Also, it is not gated `on' at index 4, because $o^<$ does not fire; (5) local gradient (output but not target), leads to potentiation of the weight of the synapses from $o^{T-}$ to $d_1$ (red), and (6) depression of $h-o$ and $d_1-o$ synaptic weights; (7) The orange spikes show the back-propagated local gradient from (1) which leads to potentiation of the $x-h$ weights. Note that for visualization purposes, the gating from $b_h$ is applied one time step later directly to $h, h^<$ and $h^>$. That is, the orange spikes in time step 7 are the full backpropagated error, but only the neurons that are also gated `on' by the combination of $h^<$ and $h^>$ are actually active in the potentiation phase in time step 8. (8) Same as (7), but the error from (2) leads to depression of the $x-h$ weights. }\label{fig:Raster}
\end{figure*}

\clearpage

\renewcommand{\arraystretch}{1.25}
\begin{table*}[!ht]
\begin{center}
{
\footnotesize
  \begin{tabular}{ | c || c | c | c | c | c | }
  \hline
  Step & 1 & 2 & 3 & 4 & 5 \\
  \hline
   & $\begin{array}{l} (x, x, g_1) \end{array}$
   & $\begin{array}{l} (m_x, x, g_{2}) \\
                       (h^<, f(W_1 x + 0.5), g_2) \\
                       (h^>, f(W_1 x - 0.5), g_2) \\
                       (h, f(W_1 x), g_2) \\ 
                       \end{array}$
   & $\begin{array}{l} (m_h, h, g_{3}) \\
                       (o^<, f(W_2 h + 0.5), g_3) \\
                       (o^>, f(W_2 h - 0.5), g_3) \\
                       (o, f(W_2 h) , g_3) \\
                       (t, t, g_3) \\
                       (b_h, h^<-h^> , g_3) \\
                     \end{array}$
  & $\begin{array}{l} (d_2^+, f(t-o)  \circ (o^<-o^>), o^<-o^>) \\ 
                      (d_2^-, f(o-t) \circ (o^<-o^>), o^<-o^>) \\
                     \end{array}$
  & $\begin{array}{l} (h, h, g_{5}) \\
                     (h^{T}, h, g_{5}) \\
                     (o, d_2^+, g_{5}) \\ 
                     (o^<, d_2^+, g_{5}) \\ 
                     (o^>, d_2^+, g_{5}) \\ 
                     (o^{T-}, d_2^-, g_{5}) \\
                     \end{array}$
  \\\hline
  & ff in & ff hid & ff out & error $ \circ$  $b_o$ & potentiation \vspace{-1ex}  \\  & & & & & $w_{hid,out}$ 
   \\\hline
  \end{tabular}}
\end{center}
\begin{center}
{
\footnotesize
  \begin{tabular}{ | c || c | c | c | c | c | c |}
  \hline
  Step & 6 & 7 & 8 & 9 & 10 & 11\\
  \hline
  & $\begin{array}{l} (h^{T}, H(W_2^T(d^+_2)\!-\!W_2^T(d^-_2)) \\
                       \!\equiv\! d_1^+, g_{6}) \\
                      \end{array}$
  & $\begin{array}{l} (x, x, g_{7}) \\
                     (h, b_h \circ d_1^+, b_h) \\
                     (h^<, b_h \circ d_1^+, b_h) \\
                     (h^>, b_h \circ d_1^+, b_h) \\
                     \end{array}$
  & $\begin{array}{l} \end{array}$
  & $\begin{array}{l} (h, h, g_{9}) \\
                     (h^{T}, h, g_{9}) \\
                     (o, d_2^-, g_{9}) \\
                     (o^<, d_2^-, g_{9}) \\ 
                     (o^>, d_2^-, g_{9}) \\ 
                     (o^{T-}, d_2^+, g_{9}) \\ 
                     \end{array}$
  & $\begin{array}{l} (h^{T}, H(-W_2^T(d_2^-)\!+\!W_2^T(d_2^+)) \\
                        \!\equiv\! d_1^-, g_{10}) \\
                     \end{array}$
  & $\begin{array}{l} (x, x, g_{11}) \\
                     (h,  b_h \circ d_1^-, b_h) \\
                     (h^<,  b_h \circ d_1^-, b_h) \\
                     (h^>,  b_h \circ d_1^-, b_h) \\
                     \end{array}$
  \\\hline
   & backpropagation & potentiation  &  & depression & backpropagation & depression 
  \\
  &  &  $w_{in,hid}$ &  &  $w_{hid,out}$ &  &  $w_{in,hid}$ \\
  \hline
  \end{tabular}}
  \caption{Information flow through the backpropagation network (see Fig.~\ref{fig:FuncConn}). Gating pulses ($g_1-g_{11}$) are sent from the gating chain. The triplet notation $(a,b,c)$ denotes information $b$ in population $a$ gated by population $c$.}
  \label{tableMNISTcircuit}
\end{center}
\end{table*}

\begin{center}
\begin{table*}[!ht]
\footnotesize
\begin{tabular}{|l|l|l|l|l|l|l|l|}
\hline
&  & \multicolumn{3}{c|}{ Power (mW)}   &   &    &  \\ \hline
& \multicolumn{1}{l|}{hardware} 
& Static &  Dynamic &  \multicolumn{1}{l|}{Total}
& \multicolumn{1}{l|}{Latency (ms)} 
& \multicolumn{1}{l|}{\begin{tabular}[c]{@{}l@{}}Dyn. Energy per \\
sample (mJ)\end{tabular}} & \begin{tabular}[c]{@{}l@{}}Energy\\delay\\product($\mu$Js)\end{tabular} \\ \hline
\multirow{3}{*}{Loihi training ds100-300-10}
& \multicolumn{1}{l|}{x86 cores}
& \Loihitrainingdownsampledpowerlakemontstatic  & \Loihitrainingdownsampledpowerlakemontdynamic & \multicolumn{1}{l|}{\Loihitrainingdownsampledpowerlakemonttotal}
& \multicolumn{1}{l|}{} & \multicolumn{1}{l|}{\LoihitrainingdownsampledenergypertrialLakemontDynamic}
& \multicolumn{1}{l|}{\LoihitrainingdownsamplededpLakemontDynamic}  \\
& \multicolumn{1}{l|}{neuron cores}
& \Loihitrainingdownsampledpowercorestatic &  \Loihitrainingdownsampledpowercoredynamic & \multicolumn{1}{l|}{\Loihitrainingdownsampledpowercoretotal}
& \multicolumn{1}{l|}{} & \multicolumn{1}{l|}{\LoihitrainingdownsampledenergypertrialLoihiDynamic}
& \multicolumn{1}{l|}{\LoihitrainingdownsamplededpLoihiDynamic}  \\
& \multicolumn{1}{l|}{whole board}
& \Loihitrainingdownsampledpowerstatic &  \Loihitrainingdownsampledpowerdynamic  & \multicolumn{1}{l|}{\Loihitrainingdownsampledpowertotal}
& \multicolumn{1}{l|}{\Loihitrainingdownsampledlatencypertrial} & \multicolumn{1}{l|}{\LoihitrainingdownsampledenergypertrialBoardDynamic} 
& \multicolumn{1}{l|}{\LoihitrainingdownsamplededpBoardDynamic} \\ \hline

\multirow{3}{*}{Loihi training 400-300-10}
& \multicolumn{1}{l|}{x86 cores}
& \Loihitrainingsmallerpowerlakemontstatic  & \Loihitrainingsmallerpowerlakemontdynamic & \multicolumn{1}{l|}{\Loihitrainingsmallerpowerlakemonttotal}
& \multicolumn{1}{l|}{} & \multicolumn{1}{l|}{\LoihitrainingsmallerenergypertrialLakemontDynamic}
& \multicolumn{1}{l|}{\LoihitrainingsmalleredpLakemontDynamic}  \\
& \multicolumn{1}{l|}{neuron cores}
& \Loihitrainingsmallerpowercorestatic &  \Loihitrainingsmallerpowercoredynamic & \multicolumn{1}{l|}{\Loihitrainingsmallerpowercoretotal}
& \multicolumn{1}{l|}{} & \multicolumn{1}{l|}{\LoihitrainingsmallerenergypertrialLoihiDynamic}
& \multicolumn{1}{l|}{\LoihitrainingsmalleredpLoihiDynamic}  \\
& \multicolumn{1}{l|}{whole board}
& \Loihitrainingsmallerpowerstatic &  \Loihitrainingsmallerpowerdynamic  & \multicolumn{1}{l|}{\Loihitrainingsmallerpowertotal}
& \multicolumn{1}{l|}{\Loihitrainingsmallerlatencypertrial} & \multicolumn{1}{l|}{\LoihitrainingsmallerenergypertrialBoardDynamic} 
& \multicolumn{1}{l|}{\LoihitrainingsmalleredpBoardDynamic} \\ \hline

\multirow{3}{*}{Loihi training 400-400-10 start}
& \multicolumn{1}{l|}{x86 cores}
& \Loihitrainingstartpowerlakemontstatic  & \Loihitrainingstartpowerlakemontdynamic & \multicolumn{1}{l|}{\Loihitrainingstartpowerlakemonttotal}
& \multicolumn{1}{l|}{} & \multicolumn{1}{l|}{\LoihitrainingstartenergypertrialLakemontDynamic}
& \multicolumn{1}{l|}{\LoihitrainingstartedpLakemontDynamic}  \\
& \multicolumn{1}{l|}{neuron cores}
& \Loihitrainingstartpowercorestatic &  \Loihitrainingstartpowercoredynamic & \multicolumn{1}{l|}{\Loihitrainingstartpowercoretotal}
& \multicolumn{1}{l|}{} & \multicolumn{1}{l|}{\LoihitrainingstartenergypertrialLoihiDynamic}
& \multicolumn{1}{l|}{\LoihitrainingstartedpLoihiDynamic}  \\
& \multicolumn{1}{l|}{whole board}
& \Loihitrainingstartpowerstatic &  \Loihitrainingstartpowerdynamic  & \multicolumn{1}{l|}{\Loihitrainingstartpowertotal}
& \multicolumn{1}{l|}{\Loihitrainingstartlatencypertrial} & \multicolumn{1}{l|}{\LoihitrainingstartenergypertrialBoardDynamic} 
& \multicolumn{1}{l|}{\LoihitrainingstartedpBoardDynamic} \\ \hline

\multirow{3}{*}{Loihi training 400-400-10 end}
& \multicolumn{1}{l|}{x86 cores}
& \Loihitrainingendpowerlakemontstatic  & \Loihitrainingendpowerlakemontdynamic & \multicolumn{1}{l|}{\Loihitrainingendpowerlakemonttotal}
& \multicolumn{1}{l|}{} & \multicolumn{1}{l|}{\LoihitrainingendenergypertrialLakemontDynamic}
& \multicolumn{1}{l|}{\LoihitrainingendedpLakemontDynamic}  \\
& \multicolumn{1}{l|}{neuron cores}
& \Loihitrainingendpowercorestatic &  \Loihitrainingendpowercoredynamic & \multicolumn{1}{l|}{\Loihitrainingendpowercoretotal}
& \multicolumn{1}{l|}{} & \multicolumn{1}{l|}{\LoihitrainingendenergypertrialLoihiDynamic}
& \multicolumn{1}{l|}{\LoihitrainingendedpLoihiDynamic}  \\
& \multicolumn{1}{l|}{whole board}
& \Loihitrainingendpowerstatic &  \Loihitrainingendpowerdynamic  & \multicolumn{1}{l|}{\Loihitrainingendpowertotal}
& \multicolumn{1}{l|}{\Loihitrainingendlatencypertrial} & \multicolumn{1}{l|}{\LoihitrainingendenergypertrialBoardDynamic} 
& \multicolumn{1}{l|}{\LoihitrainingendedpBoardDynamic} \\ \hline

\multirow{3}{*}{Loihi 400-400-10 end (learning engine off)}
& \multicolumn{1}{l|}{x86 cores}
& \Loihilearningoffpowerlakemontstatic  & \Loihilearningoffpowerlakemontdynamic & \multicolumn{1}{l|}{\Loihilearningoffpowerlakemonttotal}
& \multicolumn{1}{l|}{} & \multicolumn{1}{l|}{\LoihilearningoffenergypertrialLakemontDynamic}
& \multicolumn{1}{l|}{\LoihilearningoffedpLakemontDynamic}  \\
& \multicolumn{1}{l|}{neuron cores}
& \Loihilearningoffpowercorestatic &  \Loihilearningoffpowercoredynamic & \multicolumn{1}{l|}{\Loihilearningoffpowercoretotal}
& \multicolumn{1}{l|}{} & \multicolumn{1}{l|}{\LoihilearningoffenergypertrialLoihiDynamic}
& \multicolumn{1}{l|}{\LoihilearningoffedpLoihiDynamic}  \\
& \multicolumn{1}{l|}{whole board}
& \Loihilearningoffpowerstatic &  \Loihilearningoffpowerdynamic  & \multicolumn{1}{l|}{\Loihilearningoffpowertotal}
& \multicolumn{1}{l|}{\Loihilearningofflatencypertrial} & \multicolumn{1}{l|}{\LoihilearningoffenergypertrialBoardDynamic} 
& \multicolumn{1}{l|}{\LoihilearningoffedpBoardDynamic} \\ \hline

\multirow{3}{*}{Loihi inference net 400-400-10}
& \multicolumn{1}{l|}{x86 cores}
& \Loihiinferencepowerlakemontstatic  & \Loihiinferencepowerlakemontdynamic & \multicolumn{1}{l|}{\Loihiinferencepowerlakemonttotal}
& \multicolumn{1}{l|}{} & \multicolumn{1}{l|}{\LoihiinferenceenergypertrialLakemontDynamic}
& \multicolumn{1}{l|}{\LoihiinferenceedpLakemontDynamic}  \\
& \multicolumn{1}{l|}{neuron cores}
& \Loihiinferencepowercorestatic &  \Loihiinferencepowercoredynamic & \multicolumn{1}{l|}{\Loihiinferencepowercoretotal}
& \multicolumn{1}{l|}{} & \multicolumn{1}{l|}{\LoihiinferenceenergypertrialLoihiDynamic}
& \multicolumn{1}{l|}{\LoihiinferenceedpLoihiDynamic}  \\
& \multicolumn{1}{l|}{whole board}
& \Loihiinferencepowerstatic &  \Loihiinferencepowerdynamic  & \multicolumn{1}{l|}{\Loihiinferencepowertotal}
& \multicolumn{1}{l|}{\Loihiinferencelatencypertrial} & \multicolumn{1}{l|}{\LoihiinferenceenergypertrialBoardDynamic} 
& \multicolumn{1}{l|}{\LoihiinferenceedpBoardDynamic} \\ \hline

\multirow{3}{*}{GPU (Nvidia GTX 1660S) training} 
& \multicolumn{1}{l|}{batchsize 1}    
& 15000  & \gpuSinglepowerdynamic & \multicolumn{1}{l|}{\gpuSinglepowertotal} 
& \multicolumn{1}{l|}{\gpuSinglelatencypertrial} 
& \multicolumn{1}{l|}{\gpuSingleenergypertrialdynamic} & \multicolumn{1}{l|}{\gpuSingleedp} \\

& \multicolumn{1}{l|}{batchsize 10}
& 15000  & \gpuBatchTenpowerdynamic & \multicolumn{1}{l|}{\gpuBatchTenpowertotal} 
& \multicolumn{1}{l|}{\gpuBatchTenlatencypertrial} 
& \multicolumn{1}{l|}{\gpuBatchTenenergypertrialdynamic} & \multicolumn{1}{l|}{\gpuBatchTenedp} \\ 

& \multicolumn{1}{l|}{batchsize 100}
& 15000  & \gpuBatchHundredpowerdynamic & \multicolumn{1}{l|}{\gpuBatchHundredpowertotal } 
& \multicolumn{1}{l|}{\gpuBatchHundredlatencypertrial} 
& \multicolumn{1}{l|}{\gpuBatchHundredenergypertrialdynamic} & \multicolumn{1}{l|}{\gpuBatchHundrededp} 
\\ \hline

\end{tabular}
  \caption{Breakdown of power consumption of the chip for different cases. `ds100-300-10', means that the network is run using 100 input and 300 hidden layer neurons and with the input downsampled by 2. Here, `start' means that the measurement is taken in the first 10000 iterations of the first training epoch, and `end' means that it has been taken with the fully trained model. We use `learning engine off' to say that the weights are set to nonplastic weights after training so that the on-chip circuitry that handles learning is not active. The inference net, which uses the same weights as the fully trained network, consists only of the 3 feedforward layers and the gating chain.}
\label{tab:energy}
\end{table*}
\end{center}

\ifdefined\linenumbers \end{nolinenumbers} \fi 

\clearpage
\subsubsection{Power Measurements on GPU}
\label{gpu}
The power on the GPU was measured using \texttt{nvidia-smi} while running the TensorFlow implementation of our algorithm. The dynamic power was calculated by taking the difference between the power reading before the start of the run and during the run. The system is running Tensorflow version 2.3 on Windows 10 with an Intel i5-9600K processor and 16 GB of RAM. The GPU is an NVIDIA GeForce GTX 1600S (driver version: 461.92, CUDA version: 11.2).

\subsubsection{Detailed Description of the Algorithm}
\label{steps}

Here we go through all steps of the algorithm as shown in Fig.~\ref{fig:FuncConn}:
\begin{enumerate}
    \item MNIST images are sent to the x layer in binarized form
    \item $W_1$ is applied three times with different offsets (received from the gating chain) before application of the nonlinear activation function. This yields the hidden layer activity and start and stop learning conditions for the $W_1$ update according to Eq.~\eqref{eq:box}
    \item The MNIST labels are sent as a one-hot vector to the target layer and $W_2$ is applied three times with different offsets. This yields the output layer activity and start and stop learning conditions for the $W_2$ update according to Eq.~\eqref{eq:box}. Furthermore, the input and hidden layer activities are stored into the $m_x$ and $m_h$ relay layers for later usage in the learning phases. In the box function layer ($b_h$), the start and stop learning conditions of  $W_1$ are combined using an excitatory and inhibitory connection. This corresponds to the multiplication in Eq.~\eqref{eq:box} as in this case, multiplication is the same as taking the difference with the inverted term.
    \item The positive and negative errors are calculated by excitation from the $o$ layer and inhibition from the $t$ layer (and vice versa). The local gradient layers are not gated on by the gating chain but by the start learning condition $o^<$, and furthermore, they are inhibited by the stop learning condition $o^>$. This way, the Hadamard product from Eq.~\eqref{eq:error2} is calculated.
    \item The positive local gradient is now sent to the output layer (and its two siblings) while at the same time the hidden layer activity from time step 2 is reactivated from the relay layer. This leads to a positive weight update of $W_2$,  and because also the $d_1$ layer is activated, to updates of $W_2^T$ and $W_2^{T-}$ according to the learning rule in Eq.~\eqref{eq:learning_rule_loihi}. Weights $W_2^{T-}$ are however potentiated oppositely because layer $o^{T-}$ receives input from layer $d^-$ in this time step. This time step implements the first term of Eq.~\eqref{eq:2phaseupdate_w2}.
    \item The actual backpropagation happens in this time step.  By way of the transposed weight matrices, the $d_1$ layer receives activity from the output layer and the $o^{T-}$ layer, which contains the positive and negative local gradients. The layer is gated by the box function layer $b_h$, which calculates the Hadamard product from Eq.~\eqref{eq:error1}. This gating includes an offset so that even the smallest possible nonzero input sum leads to a spike in the neurons of the $d_1$ layer. 
    \item This time step realizes the potentiation of the $W_1$ weights. The hidden layer and its two siblings receive the local gradient that was propagated back in the previous time step, and the $x$ layer restores its previous activity from the relay layer. Because the third factor is switched on in this time step, potentiation happens according to Eq.~\eqref{eq:learning_rule_loihi}. This time step implements the first term of Eq.~\eqref{eq:2phaseupdate_w1}.
    \item To avoid interference of the input to the output layer from the previously active hidden layer (layer $h$ is connected to $o, o^<$ and $o^>$), no layer is gated on in this time step.
    \item The steps 9 to 11 are the same as 5 to 7, but with input of the opposite local gradient and without third factor active and therefore the opposite sign of learning (depression).
\end{enumerate}

\subsubsection{Synapse Types}
There are five types of synapses (see different colors in Fig.~\ref{fig:FuncConn} and Supplementary Fig.~\ref{fig:AnaConn}):
\begin{itemize}
    \item Plastic all-to-all synapses (red), that change weights according to a learning rule (Eq.~\eqref{eq:learning_rule_loihi}) and can be both positive or negative ($w_p=[-256, 254]$).
    \item Excitatory (black) one-to-one synapses with a fixed weight ($w_e = V_\mathrm{thr} = 1024$) to copy the firing pattern of the source layer to the target layer.
    \item Inhibitory (blue) one-to-one synapses with a fixed weight ($w_i=-V_\mathrm{thr} = -1024$) to always inhibit the target neurons.
    \item Excitatory one-to-all gating (green) synapses from a single gating chain neuron to all neurons of a layer with a fixed weight of usually $w_{g} = -I_{const} + 0.5 V_\mathrm{thr} = 8704$. The usual gating synapses are used to gate `on' a layer that is supposed to have the feedforward activation function $f$. Different weights $w_{g<}=-I_{const} = 8192$ and  $w_{g>}= -I_{const} + V_\mathrm{thr} = 9216$ are used to gate `on' the start and stop learning conditions to get the two step functions with which the box function is built.
    \item Excitatory one-to-one gating (yellow) synapses from the start learning layer $o^<$ and the box function layer $b_h$ to the respective local gradient layers. The $w_{g}$ weight is used for gating of the $d_2$ layers in time step 4, and the $w_{g>}$ weight is used for gating of the $d_1$ layer in time steps 6 and 10.
\end{itemize}
To a certain extent within the precision boundaries of the chip, the scaling of the weights is arbitrary. All plastic synapses and most other synapses have a synaptic delay of 0 time steps. Some synapses, specifically the ones originating from the relay layers, the $d_2$ layers, and the $b_h$ layer, have delays up to 6 time steps, i.e. their output spikes affect the target neuron several time steps later. The delays, corresponding to the horizontal arrow length, can be read from Fig.~\ref{fig:FuncConn}.

\subsubsection{Downsampled MNIST}

Table III above refers to network structures with input sizes of 400 and 100.  The 400-400-10 network used a set of trimmed images, where whitespace was removed to reduce the required size of the input layer. MNIST images for this had a border of width 4 removed. For the smaller 100-300-10 network, these images were then downsampled by averaging over $2 \times 2$ pixel grids and binarized by thresholding at 0.5.  This gave a set of $10\times 10$ binary images.

\subsubsection{Tensorflow Implementation}

For the feedforward network, our GPU tensorflow implementation is similar to the binary network presented in \cite{hubara2016binarized} but with continuous weights, and simplified to correspond with the neuromorphic design.  The loss function was modified to mean squared error and minimized with stochastic gradient descent with a fixed learning rate. Dropout and batch normalization were removed and the batch size was set to 1.

The network was initialized with Glorot parameters for each layer $l$, i.e.,  $\gamma_l = \sqrt{\frac{1.5}{N_{l-1} + N_{l}}}$  where $N_l$ is the number of neurons in layer $l$.  For each layer, weights were generated uniformly on the interval $[-\gamma_l, \gamma_l]$ and clipped to these values during learning. While these initializations do not take into account the binary activation functions of our network, tests with alternative initializations showed little or no improvement, likely due to the shallow depth of the networks considered.

Gradients on the binary network in \cite{hubara2016binarized} were calculated with a `straight-through' estimator in which the derivative of the binary activation function
\begin{equation}
    \sigma(x) = \begin{cases} 1 & x\ge 0 \\ 0 & \text{otherwise} \end{cases}
\end{equation}
was approximated as
\begin{equation}
    h_l(x) = \begin{cases}
    0 & x \ge \gamma_l \\
    \frac{1}{2\gamma_l} & -\gamma_l < x < \gamma_l \\
    0 & x < \gamma_l
    \end{cases}.
\end{equation}
This same gradient calculation is maintained, but to mimic the binary gradient signal in the neuromorphic implementation, the final gradient with respect to each weight is thresholded. If the gradient calculated with respect to weight $i$ is $g_i$, the applied gradient in the learning step is $\text{sgn}(g_i)\sigma(|g_i| - t)$ for a threshold $t$ representing the threshold of signal propagation in the neuromorphic implementation and $\sigma$ as defined above.

\newpage 
\end{widetext}

\end{document}